\documentclass{bmvc2k}

\usepackage{amsmath,amssymb} 
\usepackage{multirow}
\usepackage{adjustbox}
\usepackage{tikz}
\usepackage{comment}
\usepackage{diagbox}
\usepackage{booktabs}
\usepackage{makecell}
\usepackage{bbding}
\usepackage{hyperref}

\usepackage{color}
\newcommand{\red}[1]{\textcolor{red}{#1}}
\newcommand{\blue}[1]{\textcolor{blue}{#1}}

\title{You Only Need 90K Parameters to Adapt Light: a Light Weight Transformer for Image Enhancement and Exposure Correction}

\addauthor{Ziteng Cui}{cui@mi.t.u-tokyo.ac.jp}{1}
\addauthor{Kunchang Li}{kc.li@siat.ac.cn}{2}
\addauthor{Lin Gu}{lin.gu@riken.jp}{3,1*}
\addauthor{Shenghan Su}{su2564468850@sjtu.edu.cn}{4}
\addauthor{Peng Gao}{gaopeng@pjlab.org.cn}{2}
\addauthor{Zhengkai Jiang}{zhengkjiang@tencent.com}{5}
\addauthor{Yu Qiao}{qiaoyu@pjlab.org.cn}{2}
\addauthor{Tatsuya Harada}{harada@mi.t.u-tokyo.ac.jp}{1,3}
\addinstitution{
 The University of Tokyo
}
\addinstitution{
 Shanghai AI Laboratory
}
\addinstitution{
 RIKEN AIP
}
\addinstitution{
 Shanghai Jiao Tong University
}
\addinstitution{
 Tencent Youtu Lab
}

\runninghead{Student, Prof, Collaborator}{BMVC Author Guidelines}

\begin{document}

\maketitle

\begin{abstract}
Challenging illumination conditions (low light, under-exposure and over-exposure) in the real world not only cast an unpleasant visual appearance but also taint the computer vision tasks. After camera captures the raw-RGB data, it renders standard sRGB images with  image signal processor (ISP). By decomposing ISP pipeline into local and global image components, we propose a lightweight fast Illumination Adaptive Transformer (IAT) to restore the normal lit sRGB image from either low-light or under/over-exposure conditions. Specifically, IAT uses attention queries to represent and adjust the ISP-related parameters such as colour correction, gamma correction. With only \textbf{$\sim$90k} parameters and \textbf{$\sim$0.004s} processing speed, our IAT consistently achieves superior performance over State-of-The-Art (SOTA) on the benchmark low-light enhancement and exposure correction datasets. Competitive experimental performance also demonstrates that our IAT significantly enhances object detection and semantic segmentation tasks under various light conditions. Our code and pre-trained model is available at \href{https://github.com/cuiziteng/Illumination-Adaptive-Transformer}{this url}.

\end{abstract}

\vspace{-5mm}
\section{Introduction}
\label{sec:intro}
\begin{figure}[t]
    \centering
    \includegraphics[width=0.9\linewidth]{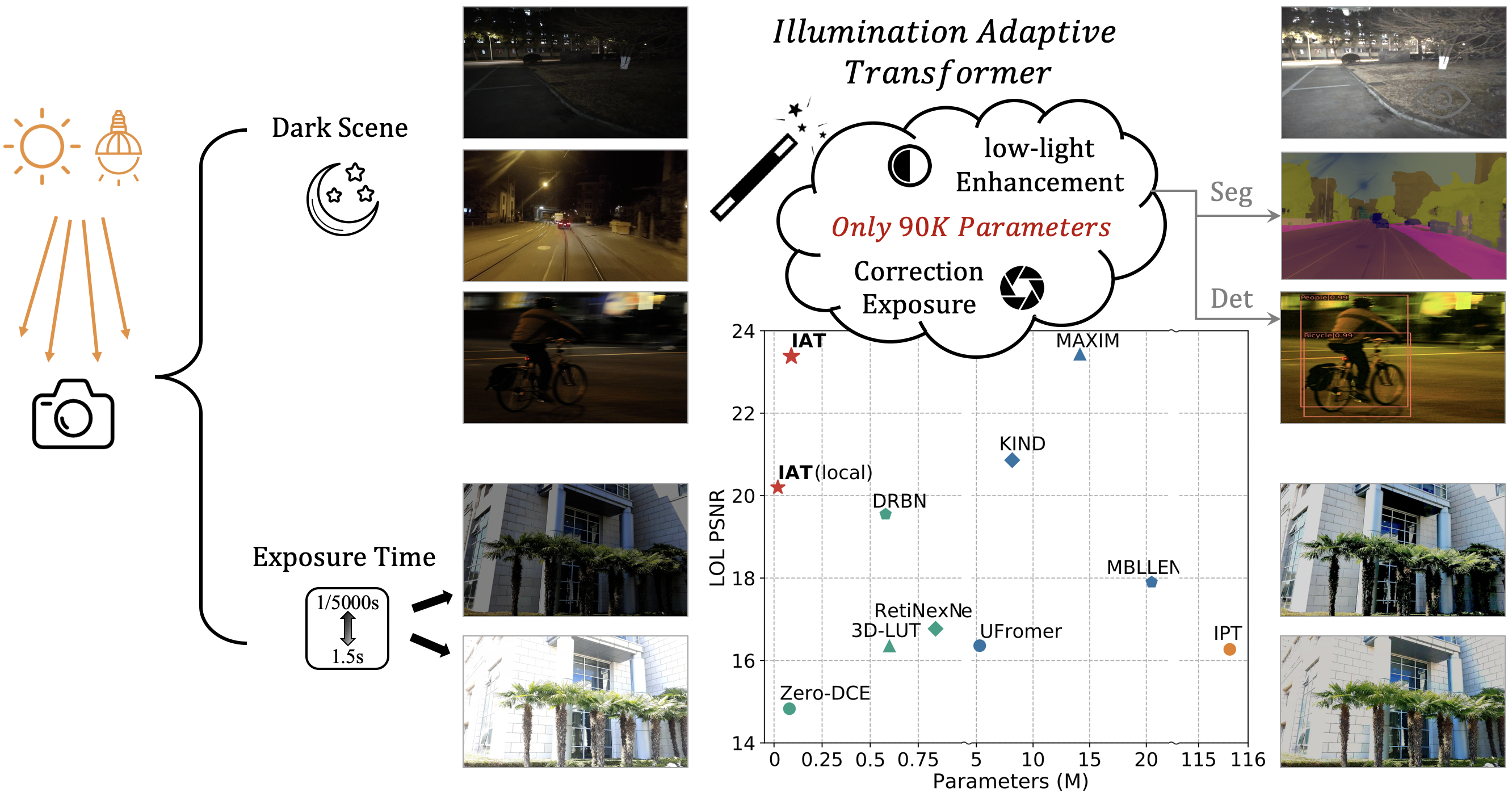}
    
    \caption{\textit{ Nature lay hid in Night}, apply our IAT, \textit{and all was light}, middle figure shows our results compared with other SOTA methods on LOL-V1 dataset~\cite{LOL_dataset}.}
    \label{fig_IAT}
    
\end{figure}

Computer vision has witnessed great success on well-taken images and videos. However, the varying light conditions in the real world poses challenges on both human visual appearance and downstream vision tasks (\textit{e.g.}, semantic segmentation and object detection). Images taken under inadequate illumination (Fig.\ref{fig_IAT} left top) suffer from limited photon counts and undesirable in-camera noise. On the other hand, outdoor scenes are often exposed to strong light such as direct sunlight (Fig.\ref{fig_IAT} left down), making image saturated due to the limited range of sensors and non-linearity in the camera image pipeline. To make it worse, both the under and over exposure may exist together, \textit{i.e.} spatial-variant illumination cast by shadow could make the contrast ratio to be 1000:1 or higher. 

Multiple techniques such as low-light enhancement~\cite{retinex,Global_HE,LIME,LLNet,Lv2018MBLLEN,zero_dce,KIND,RCT_ICCV21,3DLUT,Deep_LPF,MLP_enhancement,Tan_enhancement}, exposure correction~\cite{exposure_ICCV2003,Exposure_ECCV12,Exposure_2021_CVPR,Learn_adapt_light,exposure_bmvc21} have been proposed to adapt to the difficult light condition. Low-light enhancement methods restore the details while suppressing the accompanying noises. Exposure correction methods focus on adjusting the under/over exposure image to reconstruct a clear image against short/long exposure time. While many efforts elaborate on improving human oriented visual perceptual, there also several methods  enhance the high-level tasks by boosting their robustness against low light~\cite{ICCV_MAET,IJCV_21_lowface,Hong_2021_BMVC,yolo-in-the-dark} and over-exposure conditions~\cite{DBGAN_2021_WACV_strong_light}. As shown in Fig.\ref{fig_IAT}, we aim for a unified lightweight framework that  improves both the visual appearance and consequent recognition tasks under challenging real-world light condition. 


While sRGB images are most common to everyday life, many existing light adaptive methods operate on raw-RGB images which linearly proportion to the actual scene irradiance. To directly process the sRGB images, we specifically considers the image signal processor (ISP) in camera that renders sRGB from raw-RGB image. We propose a novel two-branches transformer based model to handle this issue, a pixel-wise local branch $f$ coupled with global ISP branch $g$. In local branch $f$, we map the input image to latent feature space and replace transformer's attention block to depth-wise convolution for light-weight design. In global branch $g$, we use transformer's attention queries to control and adjust the global ISP-related parameters (\textit{i.e.} colour transform matrix, gamma value). In addition, the learned queries could dynamic change under different light condition at same time (\textit{i.e.} over-exposure and under exposure).

Extensive experiments are conducted on several real-world and synthetic datasets, \textit{i.e.}, image enhancement dataset LOL~\cite{LOL_dataset} and photo retouching dataset MIT-Adobe FiveK~\cite{fivek_dataset}, low-light detection dataset EXDark~\cite{EXDark} and low-light segmentation dataset ACDC~\cite{ACDC}. Results show that our IAT achieve state-of-the-art performance across a range of low-level and high-level tasks. More importantly, our IAT model is of only \textbf{0.09M} parameters, much smaller than current SOTA transformer-based models~\cite{IPT_CVPR,Wang2022Uformer,MLP_enhancement}. Besides, our average inference speed is \textbf{0.004s} per image, faster than the SOTA methods taking around 1s per image.

Our contribution could be summarised as follow:
\vspace{-2mm}
\begin{itemize}
    \item We have proposed a fast and light-weight framework, Illumination Adaptive Transformer (IAT), to adapt to challenging light conditions in the real world, which could both handle the low-light enhancement and exposure correction tasks.
    \vspace{-1mm}
    \item We have proposed a novel transformer-style structure to estimate ISP-related parameters to fuse the target sRGB image, wherein the learnable attention quires are utilised to attend the whole image, also we replace the layer norm to a new light normalisation, for better handling low-level vision tasks.
    \vspace{-1mm}
    \item Extensive experiments on several real-world datasets on 3 low-level tasks and 3 high-level tasks demonstrate the superior performance of IAT over SOTA methods. IAT is light weight and  mobile-friendly with only \textbf{0.09M} model parameters and \textbf{0.004s} processing time per image. We will release the source code upon publication.
\end{itemize}
\vspace{-2mm}

\section{Related Works}

\label{sec:related_works}
\subsection{Enhancement against Challenging Light Condition}

\noindent
Earlier low-light image enhancement solutions mainly rely on RetiNex theory~\cite{retinex} or histogram equalization~\cite{DIP,Global_HE}. Since LLNet~\cite{LLNet} utilised a deep-autoencoder structure, CNN based methods~\cite{Lv2018MBLLEN,KIND,RCT_ICCV21,zero_dce,Deep_LPF,DeepUPE_2019_CVPR,fu2022gan,Learn_adapt_light,SNR_2022_CVPR,Tan_enhancement} have been widely used in this task and gain SOTA results on the benchmark enhancement datasets~\cite{fivek_dataset,LOL_dataset}.

Similar to low-light enhancement, traditional exposure
correction algorithms~\cite{exposure_ICCV2003,Exposure_ECCV12} also use image histograms to adjust image intensities. The strategy then tends to correct exposure errors by adjusting the tone curve via a trained deep learning model~\cite{YuNIPS18,park2018distort}. Very recently, Afifi \textit{et al.}~\cite{Exposure_2021_CVPR} propose a coarse-to-fine neural network to correct photo exposure, after that Nsampi \textit{et al.}~\cite{exposure_bmvc21} introduce attention mechanism into this task. 

Beyond low-level vision, low-light/ strong-light scenario also deteriorates the performance of high level vision ~\cite{ICCV_MAET,Hong_2021_BMVC,IJCV_21_lowface,Optical_Flow_in_dark,DBGAN_2021_WACV_strong_light,yolo-in-the-dark}. Several methods based on data synthesis~\cite{Optical_Flow_in_dark}, self-supervised learning~\cite{ICCV_MAET} and domain adaptation~\cite{yolo-in-the-dark} have been proposed to support high level vision tasks under challenging illumination conditions.

\vspace{-1mm}
\subsection{Vision Transformers}
Transformer~\cite{Transformer_NLP} was firstly proposed in NLP area to capture long-range dependencies by global attention. ViT~\cite{ViT} made the first attempt in vision task by splitting the image into tokens before sending into transformer model. Since then, Transformer based models have gained superior performances in many computer vision tasks, including image/video classification~\cite{swin,uniformer1}, object detection~\cite{DETR,zhang2022monodetr}, semantic segmentation~\cite{xie2021segformer}, vision-language model~\cite{clip,zhang2021tip} and so on.

In low-level vision area, transformer-based models has also made much progress on several sub-directions, such as image super-resolution~\cite{liang2021swinir}, image restoration~\cite{zamir2021restormer,Wang2022Uformer,IPT_CVPR}, image colorization~\cite{kumar2021colorization} and bad weather restoration~\cite{TransWeather}. 
Very recently, MAXIM~\cite{MLP_enhancement} use MLP-based model~\cite{mlpmixer} in low-level vision area which also shows MLP's potential on low-level vision tasks. However, existing transformer $\&$ MLP models require much computational cost (\textit{e.g.} 115.63M for IPT~\cite{IPT_CVPR}, 14.14M for MAXIM~\cite{MLP_enhancement}), making it hard to implement on mobile and edge devices. Extreme lightweight of our method (0.09M) is particular important in low-level vision and computational photography.


\vspace{-1mm}
\section{Illumination Adaptive Transformer}
\vspace{-1mm}
\subsection{Motivation}
\label{sec:physic_ana}

For a sRGB image $I_i$ taken from light condition $L_i$, the input photons under light condition $L_i$ would project through the lens on capacitor cluster, to pass by the in-camera process~\cite{CVPRLow_light_denoise_physics} and render with image signal processor (ISP) pipeline $G(\cdot)$~\cite{ISP_pipeline_eccv16,brooks2019unprocessing}. Our goal is to match input sRGB $I_i$  to the target sRGB image $I_t$ (taken under light condition $L_t$). Existing deep-learning based methods tend to build an end-to-end mapping between $I_i$ and $I_t$~\cite{LLNet,Lv2018MBLLEN,Exposure_2021_CVPR} or estimate some high-level representation to assist enhancement task (\textit{i.e.} illumination map~\cite{DeepUPE_2019_CVPR}, colour transform function~\cite{RCT_ICCV21}, 3D look-up table~\cite{3DLUT}). However, the actual lightness degradation happens in raw-RGB space, and the processes in camera ISP involves more elaborated non-linear operations such as white balance, colour space transform, gamma correction, \textit{etc.}
Therefore, much of research conducts image enhancement~\cite{see_in_the_dark,CVPRLow_light_denoise_physics} directly on raw-RGB data rather than sRGB images.

To this end, Brooks \textit{et al.}~\cite{brooks2019unprocessing} inverse each steps in ISP pipeline (\textit{i.e.} gamma correction, tone mapping, camera colour transformation) to transform input sRGB image to "unprocessed" raw-RGB data. After that, Afifi and Brown~\cite{Afifi_2020_CVPR} apply an encoder-decoder structure to edit the illumination of sRGB image from input light $I_i$ to target light $I_t$ as following:

\begin{equation}
    I_t = G(F(I_i)),
\label{eq:simple_Function_P}
\end{equation}
where $F$ is an unknown reconstruction function maps $I_i$ to the corresponding raw-RGB data $D = F(I_i)$, and $G$ is camera rendering function that transform $D$ back to target sRGB image $I_t$. Here~\cite{Afifi_2020_CVPR} use the network encoder $f$ to represent $F$, before adding several individual decoders $g_t$ upon encoder $f$. The function  maps  $f(I_i)$ to target $I_t$ illumination conditions is represented below:

\begin{equation}
    I_t = g_t(f(I_i)),
\label{eq:simple_Function}
\end{equation}

 For the sake of lightweight network design, inspired by the DETR~\cite{DETR} which controls different object proposals via transformer queries, here we use different queries to control the ISP-related parameters in $g_t(\cdot)$.  This re-configures parameters to make the image $I_i$ adaptive to target light condition $L_t$. In training stage, the queries is dynamically updated in each iteration to match the target image $I_t$. 
Here we simplify the ISP procedures~\cite{brooks2019unprocessing,ICCV_MAET,ISP_uprocess_2021_ICCV} into the equation~\ref{eq:ISP_Function_detail} below. The simplification details could be found in the supplementary.


\begin{equation}
g_t(\cdot) = (max(\sum_{c_j} W_{c_i,c_j} (\cdot), \epsilon))^\gamma,  c_i,c_j \in \{r,g,b\}.
\label{eq:ISP_Function_detail}
\end{equation}

 $W_{c_i,c_j}$ is a $3 \times 3$ joint colour transformation matrix, considering the white balance and colour transform matrix. We adopt 9 queries to control $W_{c_i,c_j}$'s parameters. $\gamma$ denotes the gamma correction parameter which we use a single query to control. $\epsilon$ is a very small value to prevent numerical instability. Here we set $\epsilon = 1e^{-8}$ in our experiments.

For process $F$, we apply a pixel-wise least squares model $f$. Our $f$
consists of two individual branches to predict multiply map $M$ and add map $A$. We then apply a least squares to process input sRGB image: $f(I_i) = I_i\odot M + A$. Here $M$ and $A$ has the same size with $I_i$ to complete pixel-level multiplicative and additive adjustment. Finally, the equation of our IAT model follows:

\begin{equation}
    I_t = (max(\sum_{c_j} W_{c_i,c_j}(I_i\odot M + A)), 0)^\gamma.
\label{eq:total_function}
\end{equation}

The non-linear operations are decomposed into a local pixel-wise components $f$ and a global ISP components $g$. Thus, we design two individual transformer style branches: local adjustment branch and global ISP branch, to estimate the local pixel-wise components and global ISP components respectively.

\begin{figure}[t]
    \centering
    \includegraphics[width=1.0\linewidth]{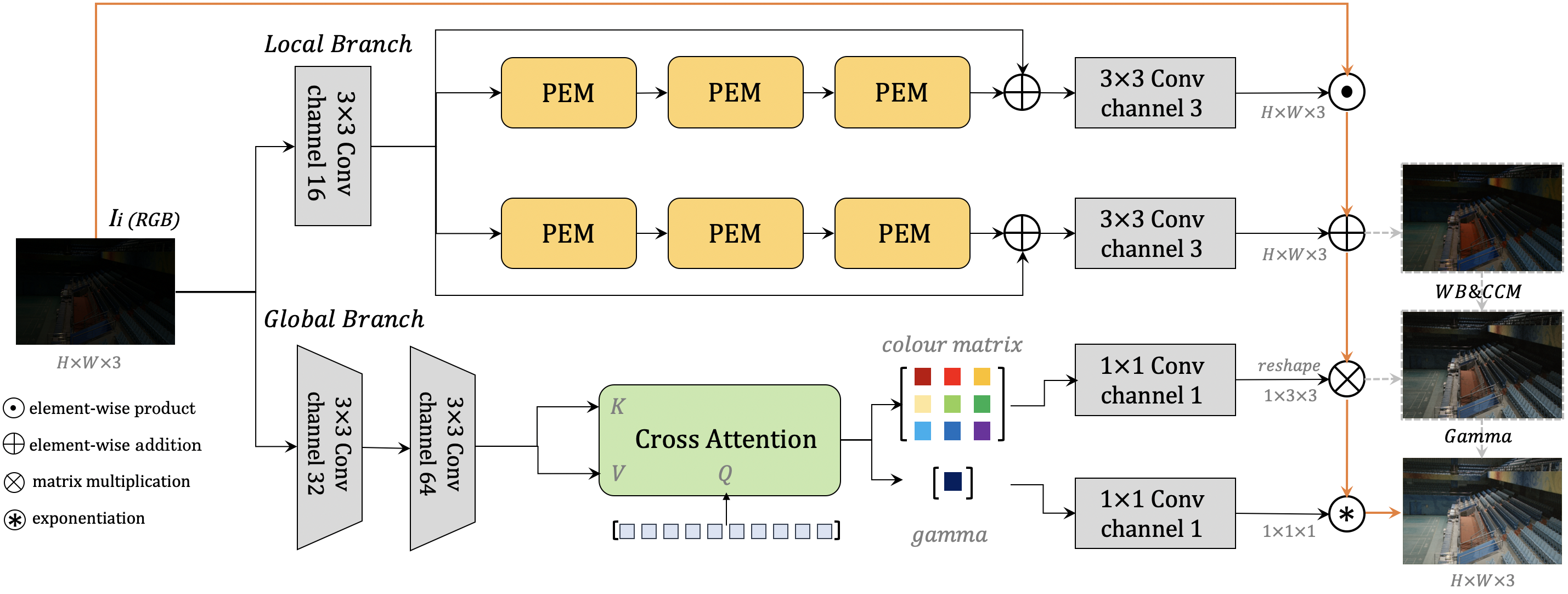}
    
    \caption{Structure of our Illumination Adaptive Transformer (IAT), the black line refers to the parameters generation while the yellow line refers to image processing.}
    \label{fig_model_structure}
    
\end{figure}

\subsection{Model Structure}
\label{sec:model}
Given an input sRGB image $I_i \in \mathbb{R}^{H \times W \times 3}$ under light condition $L_i$, where $H \times W$ denotes the size dimension and $3$ denotes the channel dimension ($\{r,g,b\}$). As shown in Fig.\ref{fig_model_structure}, we propose our Illumination Adaptive Transformer (IAT) to transfer the input RGB image  $I_i$  to a target RGB $I_t \in \mathbb{R}^{H \times W \times 3}$ under the proper uniform light $L_t$. 


\textbf{Local Branch.}
In the local branch, we focus on estimating the local components $M, A$ to correct the effect of illumination. Instead of adopting a U-Net~\cite{unet} style structure, which downsamples the images first before upsampling, we aim to maintain the input resolution through the local branch to preserve the informative details.  Therefore, we propose a transformer-style architecture for the local branch. Compared to popular U-Net style structures~\cite{Lv2018MBLLEN,Exposure_2021_CVPR}, our structure could deal with arbitrary resolution images without resizing them.

\begin{figure}[t]
    \centering
    \includegraphics[width=1.0\linewidth]{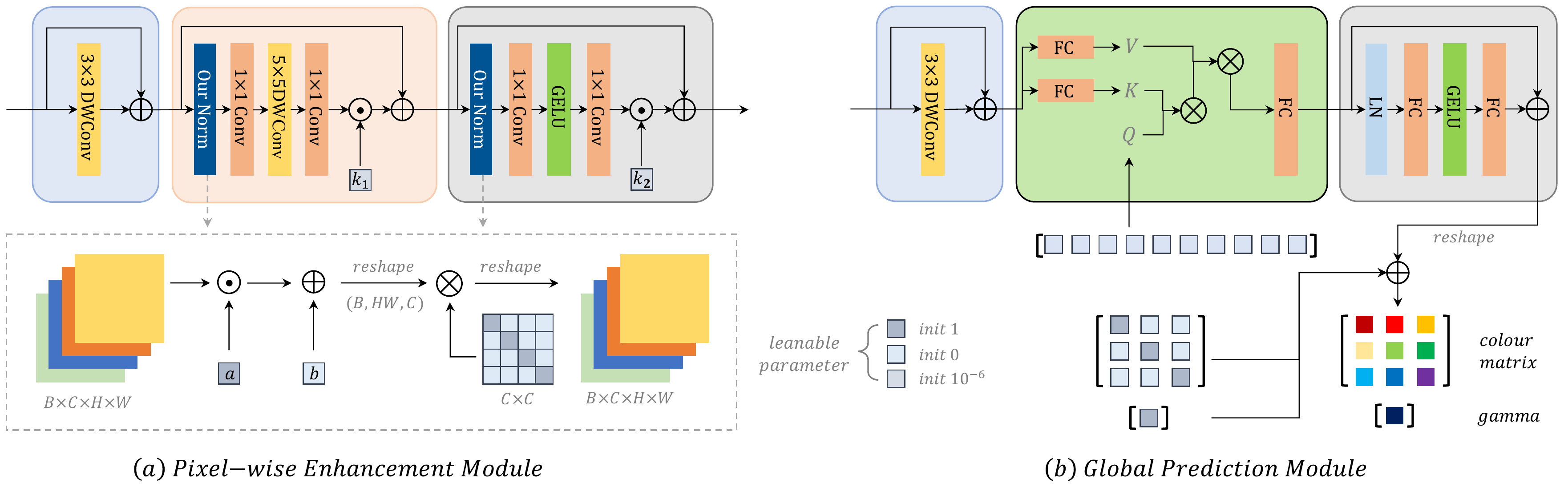}
    
    \caption{Detailed structure of Pixel-wise Enhancement Module (PEM) and Global Prediction Module (GPM).}
    \label{detailed_structure}
    
\end{figure}

At first, we expand the channel dimension via a 3$\times$3 convolution and pass them to two independent branches stacked by Pixel-wise Enhancement Module (PEM). For the lightweight design in the local branch, we replace self-attention with depth-wise convolution as suggested in the previous works~\cite{han2021connection,uniformer1}, depth-wise convolution could reduce parameters and further save computation cost.
As shown in Fig.~\ref{detailed_structure} (a), our PEM first encodes the position information by 3$\times$3 depth-wise convolution before enhancing local details with  PWConv-DWConv-PWConv. Finally, we adopt two 1$\times$1 convolutions to enhance token representation individually.
Specially, we design new kind of normalisation names light normalisation, to replace transformer's Layer Normalisation~\cite{ln}. As shown in Fig.~\ref{detailed_structure} (a), light normalisation learns to scale $a$ and bias $b$ via two learnable parameters before fusing the channels via the learnable matrix. The matrix is initialised as an identity matrix.
For better convergence, we adopt Layer Scale~\cite{cait} which multiplies the features by a small number $k_1$/$ k_2$.

We stack 3 PEMs in each branch and then connect the output features with the input features through element-wise addition. This skip connection~\cite{ResNet} helps maintain the original image details. Finally,
we decrease the channel dimension by a 3$\times$3 convolutions and adopt ReLU/ Tanh function to generate the local components $M$/ $A$ in Eq.~\ref{eq:total_function}.

\textbf{Global ISP Branch.} 
Global ISP branch accounts for part of the ISP pipeline~\cite{ISP_flexisp,ISP_L3method,ISP_pipeline_eccv16,brooks2019unprocessing} (\textit{i.e.} gamma correction, colour matrix transform, white balance) when transferring the target RGB image $I_t$. Specifically, the value of each pixel in the target image is determined by a global operation defined in Eq.\ref{eq:ISP_Function_detail}.

Inspired by Detection Transformer DETR~\cite{DETR},  we design global component queries to decode and predict the  $W, \gamma$ to generate sRGB image $I_t$. This transformer structure allows capturing global interactions between context and individual pixels. As shown in Fig.~\ref{fig_model_structure}, we first stack two convolutions as a lightweight encoder,
which encodes the features in a high dimension with lower resolution, on the one hand lower resolution would save computational cost which contribute to the light-weight design, on the other hand higher feature representation would be helpful to extract image's global-level features. Then the generated features are passed to the Global Prediction Module (GPM), Fig.~\ref{detailed_structure} (b) shows the detailed structure of GPM, different from original DETR~\cite{DETR} model,
our global component queries Q are initialised as zeros without extra multi-head self-attention. Q is global component learnable embedding that attends keys K and values V generated from encoded features. The positional encoding for K and V is from a depth-wise convolution, which is friendly with different input resolutions. After feed forward network (FFN)~\cite{ViT} with two linear layers, we add two extra parameters with special initialisation to output colour matrix and gamma. This initialisation makes sure the colour matrix is identity matrix $W$ and the gamma value $g$ is one in the beginning, thus contributing to stable training.

\vspace{-2mm}
\section{Experiments}
\vspace{-1mm}
We evaluate our proposed IAT model on benchmark datasets and experimental settings for both low-level and high-level vision tasks under different illumination conditions. Three low-level vision tasks include: $(a)$ image enhancement (LOL (V1 \& V2-real)~\cite{LOL_dataset}), $(b)$ image enhancement (MIT-Adobe FiveK~\cite{fivek_dataset}), $(c)$ exposure correction~\cite{Exposure_2021_CVPR}. Three high-level visions tasks include: $(d)$ low-light object detection $(e)$ low-light semantic segmentation $(f)$ various-light object detection. The number of PEM number in local branch are both set to 3, while the channel number in PEM is set to 16. 

For all low-level vision experiments: $\left\{ (a), (b), (c) \right\}$, the IAT model are trained on a single GeForce RTX 3090 GPU with batch size 8. We use Adam optimizer to train our IAT model while the initial learning rate and weight decay are separately set to $2e^{-4}$ and $1e^{-4}$. A cosine learning schedule has also been adopted to avoid over-fitting. For data augmentation, horizontal and vertical flips have been used to acquire better results.

\begin{figure}[t]
    \centering
    \includegraphics[width=0.9\linewidth]{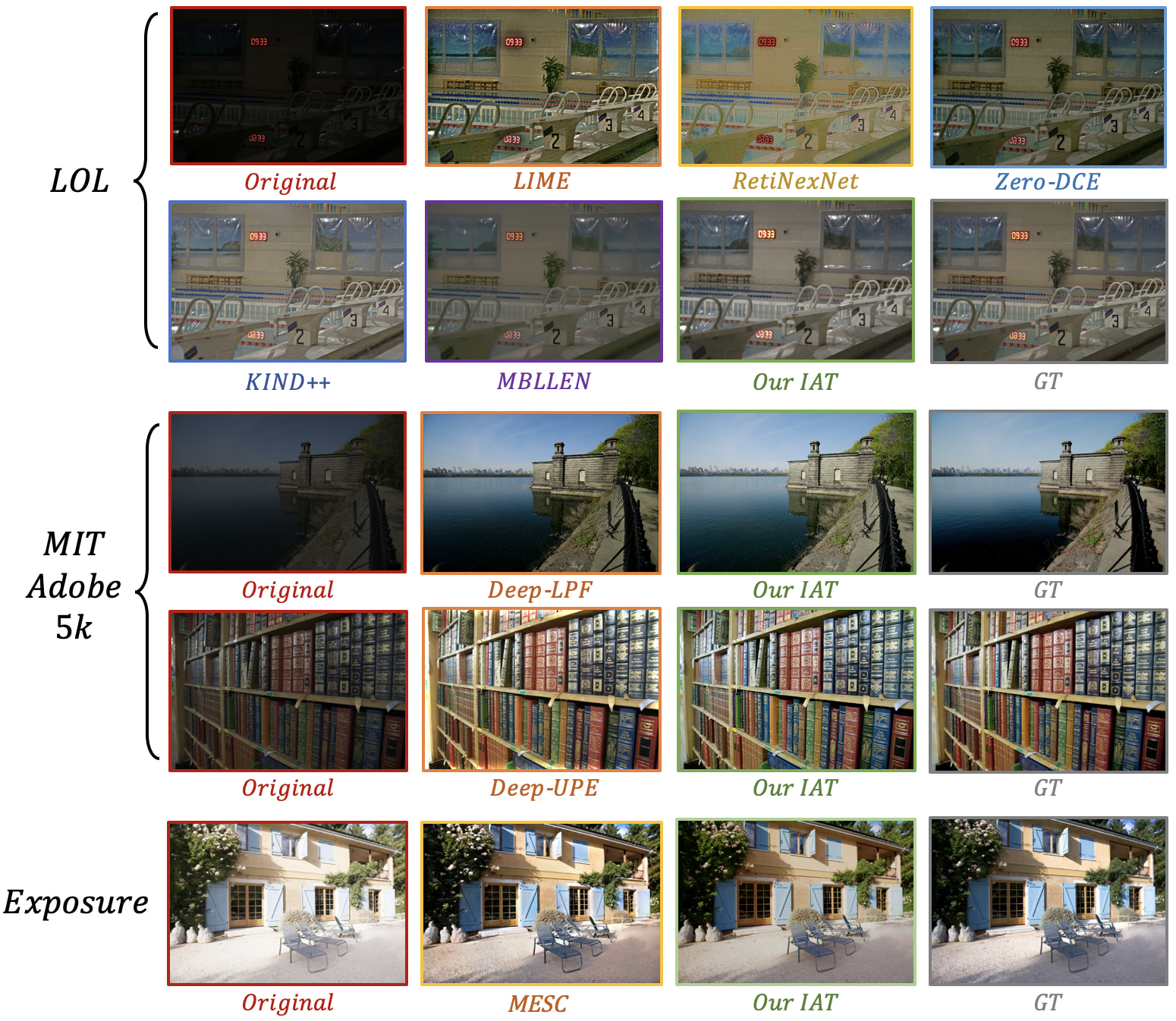}
    \caption{Results on enhancement dataset \cite{LOL_dataset,fivek_dataset}) and exposure correction dataset~\cite{Exposure_2021_CVPR}.}
    \label{fig:low_level}
    
\end{figure}

\subsection{Low-level Image Enhancement.}
\label{exp:image_enhancement}
For $(a)$ and $(b)$ image enhancement task, we evaluate our IAT framework on benchmark datasets: LOL (V1 \& V2-real)~\cite{LOL_dataset} and MIT-Adobe FiveK~\cite{fivek_dataset}. 

LOL~\cite{LOL_dataset} has two versions: LOL-V1 consists of 500 paired normal-light images and low-light images. 485 pairs are used for training and the other 15 pairs are for testing.
LOL-V2-real consists of 789 paired normal-light images and low-light pairs. 689 pairs are used for training and the other 100 pairs are for testing. The loss function between input image $I_i$ and target image $I_t$ for LOL dataset training is a mixed loss function~\cite{TransWeather}  consisting of smooth L1 loss and VGG loss~\cite{Perceptual_loss}. In LOL-V1 training, the images are cropped into $256 \times 256$ to train 200 epochs and then fine-tune on $600 \times 400$ resolution for 100 epochs. In LOL-V2-real training, the image resolution is maintained at $600 \times 400$ and trained for 200 epochs. Both LOL-V1 and LOL-V2-real testing the image resolution is maintained at $600 \times 400$. We compare our method with SOTA methods~\cite{LIME,LOL_dataset,Lv2018MBLLEN,KIND,zero_dce,DRBN_CVPR,RCT_ICCV21,IPT_CVPR,3DLUT,Wang2022Uformer,MLP_enhancement}. For image quality analysis, we evaluate the peak signal-to-noise ratio (PSNR) and structural
similarity index measure (SSIM). For the model efficiency analyze, we report three metrics: FLOPs, model parameters and test time, as shown in the last column of Table.\ref{tab:LOL}. We list different model's test time on their corresponding code platform (M means Matlab, T means TensorFlow, P means PyTorch). As shown in Table~\ref{tab:LOL}, \textbf{IAT (local)} only uses the local network to train the model and \textbf{IAT} refers to the whole framework. Our \textbf{IAT} gains SOTA result on both image quality and model efficiency, especially less than 100$\times$ FLOPs and parameters usage compare to the current SOTA methods MAXIM~\cite{MLP_enhancement}.

MIT-Adobe FiveK~\cite{fivek_dataset} dataset contains 5000 images, each  was manually enhanced by five different experts (A/B/C/D/E). Following the previous settings~\cite{DeepUPE_2019_CVPR,Deep_LPF}, we only use experts C's adjusted images as ground truth images. For MIT-Adobe FiveK~\cite{fivek_dataset} dataset training, we use a single L1 loss function to optimize IAT model. Our method is compared with SOTA enhancement methods~\cite{white_box,unet,DPE_CVPR18,DPED,DeepUPE_2019_CVPR,DeepUPE_2019_CVPR,Deep_LPF,3DLUT} on FiveK dataset. The image quality results (PSNR, SSIM) and model parameters are reported in Table.~\ref{tab:MIT5k}. Our \textbf{IAT} also gain satisfactory result in both quality and efficiency. Qualitative results of LOL~\cite{LOL_dataset} and FiveK~\cite{fivek_dataset} has been shown in Fig.\ref{fig:low_level}. More results could be found in supplementary material.  

\begin{table}[t]
\caption{Experimental results on LOL (V1 \& V2)~\cite{LOL_dataset} datasets, best and second best results are marked in red and blue respectively, noted here~\cite{LIME} is non-deep learning method and~\cite{zero_dce} is self-supervised learning method.}\label{tab:LOL}
\centering
\setlength\tabcolsep{4pt}
\begin{adjustbox}{max width = 0.95\linewidth}
\begin{tabular}{l|cc|cc|ccc}
\Xhline{1.0pt}
\multirow{2}{*}{Methods}         & \multicolumn{2}{c|}{LOL-V1} & \multicolumn{2}{c|}{LOL-V2-real} & \multicolumn{3}{c}{Efficiency}             \\ \cline{2-8}  & PSNR$\uparrow$ & SSIM$\uparrow$ & PSNR$\uparrow$ & SSIM$\uparrow$   & FLOPs(G)$\downarrow$ & \#Params(M)$\downarrow$ & test time(s)$\downarrow$ \\ \Xhline{0.6pt}
LIME*~\cite{LIME}                  & 16.67        & 0.560        & 15.24        & 0.470        & -         & -              & 3.241 (M)     \\
Zero-DCE*~\cite{zero_dce}    & 14.83        & 0.531        & 14.32        & 0.511        & 2.53      & \blue{0.08}           & \red{0.002} (P)     \\
RetiNexNet~\cite{LOL_dataset}   & 16.77        & 0.562        & 18.37        & 0.723        & 587.47    & 0.84           & 0.841 (T)     \\
MBLLEN~\cite{Lv2018MBLLEN}  & 17.90        & 0.702        & 18.00        & 0.715        & 19.95     & 20.47          & 1.981 (T)     \\
DRBN~\cite{DRBN_CVPR}    & 19.55        & 0.746        & 20.13        & \blue{0.820}        & 37.79     & 0.58           & 1.210 (P)     \\
3D-LUT~\cite{3DLUT}     & 16.35        & 0.585        & 17.59        & 0.721        & 7.67      & 0.6            & 0.006 (P)     \\
KIND~\cite{KIND}   & 20.86        & 0.790        & 19.74        & 0.761        & 356.72    & 8.16           & 0e38 (T)     \\
UFormer~\cite{Wang2022Uformer}      & 16.36        & 0.771        & 18.82        & 0.771        & 12.00     & 5.29           & 0.248 (P)     \\
IPT~\cite{IPT_CVPR}      & 16.27        & 0.504        & 19.80        & 0.813        & 2087.35   & 115.63         & 1.365 (P)     \\
RCT~\cite{RCT_ICCV21}       & 22.67       & 0.788        & -            & -            & -         & -              & -  \\       
MAXIM~\cite{MLP_enhancement}        & \red{23.43}     & \red{0.863} & \blue{22.86}     & 0.818       & 216.00  & 14.14    & 0.602 (P) \\ \Xhline{0.6pt}
\multicolumn{1}{l|}{\textbf{IAT (local)}} & 20.20        & 0.782        & 20.30        & 0.789        & \red{1.31}      & \red{0.02}           & \red{0.002} (P)        \\
\textbf{IAT }      &  \blue{23.38}  &   \blue{0.809}   & \red{23.50}       & \red{0.824}        & \blue{1.44}      & 0.09           & \blue{0.004} (P)         \\ \Xhline{1.0pt}
\end{tabular}
\end{adjustbox}
\end{table}

\begin{table}[t]
\caption{Experimental results on MIT-Adobe FiveK~\cite{fivek_dataset} dataset.}
\centering
\setlength\tabcolsep{2pt}
\begin{adjustbox}{max width = 0.98\linewidth}
\begin{tabular}{c|ccccccccc}
\Xhline{1.0pt}
Metric     & White-Box~\cite{white_box}  & U-Net~\cite{unet}  & DPE~\cite{DPE_CVPR18}   & DPED~\cite{DPED}  & D-UPE~\cite{DeepUPE_2019_CVPR} & D-LPF~\cite{Deep_LPF} & 3D LUT~\cite{3DLUT}  & \textbf{IAT}   \\ \hline
PSNR$\uparrow$       & 18.57   & 21.57 & 23.80 & 21.76 & 23.04    & 23.63    & 25.21 & \textbf{25.32} \\ \hline
SSIM$\uparrow$       & 0.701    & 0.843 & 0.880 & 0.871 & 0.893    & 0.875    & \textbf{0.922} & 0.920 \\ \hline
\#Params.$\downarrow$ & -     & 1.3M  & 3.3M  & -     & 1.0M     & 0.8M     & 0.6M & \textbf{0.09M} \\
\Xhline{1.0pt}
\end{tabular}
\end{adjustbox}
\label{tab:MIT5k}
\vspace{-3mm}
\end{table}

\begin{table}[t]
\caption{Experimental results on exposure correction dataset~\cite{Exposure_2021_CVPR}. Note here HE and LIME~\cite{LIME} are non-deep learning methods. PSNR, SSIM  and PI results, reported by competing works, are from~\cite{Exposure_2021_CVPR}.}
\centering
\setlength\tabcolsep{2pt}
\begin{adjustbox}{max width = 1\linewidth}
\begin{tabular}{l|cc|cc|cc|cc|cc|cc|c}
\Xhline{1.0pt}
\multirow{2}{*}{Method} & \multicolumn{2}{c|}{Expert A} & \multicolumn{2}{c|}{Expert B} & \multicolumn{2}{c|}{Expert C} & \multicolumn{2}{c|}{Expert D} & \multicolumn{2}{c|}{Expert E}      & \multicolumn{2}{c|}{Avg} & \multirow{2}{*}{PI$\downarrow$} \\ \cline{2-13}
& PSNR$\uparrow$ & SSIM$\uparrow$  & PSNR$\uparrow$ & SSIM$\uparrow$          & PSNR$\uparrow$ & SSIM$\uparrow$   & PSNR$\uparrow$   & SSIM$\uparrow$   & PSNR$\uparrow$  & SSIM$\uparrow$ & PSNR$\uparrow$& SSIM$\uparrow$  & \\
\Xhline{0.6pt}
HE*~\cite{DIP}                      & 16.14         & 0.685         & 16.28         & 0.671         & 16.52         & 0.696         & 16.63         & 0.668         & 17.30 & 0.688 & 16.58       & 0.682      & 2.405               \\
LIME*~\cite{LIME}                    & 11.15         & 0.590          & 11.83         & 0.610         & 11.52         & 0.607         & 12.64         & 0.628         & 13.61 & 0.653 & 12.15       & 0.618      & 2.432               \\
DPED~\cite{DPED} (Sony)             & 17.42         & 0.675         & 18.64         & 0.701         & 18.02         & 0.683         & 17.55         & 0.660         & 17.78 & 0.663 & 17.88       & 0.676      & 2.806               \\
DPE~\cite{DPE_CVPR18} (S-FiveK)           & 16.93         & 0.678         & 17.70         & 0.668         & 17.74         & 0.696         & 17.57         & 0.674         & 17.60 & 0.670 & 17.51       & 0.677      & 2.621               \\
RetinexNet~\cite{LOL_dataset}              & 10.76         & 0.585         & 11.61         & 0.596         & 11.13         & 0.605         & 11.99         & 0.615         & 12.67 & 0.636 & 11.63       & 0.607      & 3.105               \\
Deep-UPE~\cite{DeepUPE_2019_CVPR}                & 13.16         & 0.610         & 13.90         & 0.642         & 13.69         & 0.632         & 14.80         & 0.649         & 15.68 & 0.667 & 14.25       & 0.640      & 2.405               \\ 
Zero-DCE~\cite{zero_dce}                & 11.64         & 0.536         & 12.56         & 0.539         & 12.06         & 0.544         & 12.96         & 0.548         & 13.77 & 0.580 & 12.60       & 0.549      & 2.865               \\ 
MSEC~\cite{Exposure_2021_CVPR}                    & 19.16         & 0.746         & 20.10         & 0.734         & 20.20         & 0.769         & 18.98         & 0.719         & 18.98 & 0.727 & 19.48       & 0.739      & 2.251              \\ 
\Xhline{0.6pt}
\textbf{IAT (local) }            &     16.61    &      0.750   &    17.52    &  0.822  &  16.95  &  0.780  &  17.02  &  0.773  & 16.43  & 0.789  & 16.91 &  0.783  &   2.401           \\ 
\textbf{IAT}                     & \textbf{19.90 }        & \textbf{0.817}         & \textbf{21.65 }        & \textbf{0.867}         & \textbf{21.23}         & \textbf{0.850}  & \textbf{19.86}   & \textbf{0.844}    & \textbf{19.34} & \textbf{0.840} & \textbf{20.34}       & \textbf{0.844}      & \textbf{2.249}               \\ 
\Xhline{1.0pt}
\end{tabular}
\end{adjustbox}
\label{tab:Exposure}
\end{table}

\vspace{-3mm}
\subsection{Exposure Correction.}
\label{exp:exposure_correction}
For the $(c)$ exposure correction task, we evaluate IAT on the benchmark dataset proposed by~\cite{Exposure_2021_CVPR}. The dataset contains 24,330 sRGB images, divided into 17,675 training images, 750 validation images, and 5905 test images. Images in~\cite{Exposure_2021_CVPR} are adjusted by MIT-Adobe FiveK~\cite{fivek_dataset} dataset with 5 different exposure values (EV), ranging from under-exposure to over-exposure condition. Same as~\cite{fivek_dataset}, test set has 5 different experts' adjust results (A/B/C/D/E). Following the setting of~\cite{Exposure_2021_CVPR}, the training images are cropped to $512 \times 512$ patches and the test image is resized to have a maximum dimension of $512$ pixels. We compare the test images with all five experts' results. Here we use L1 loss function for exposure correction training. 

The evaluation result is shown in Table.~\ref{tab:Exposure}. Our comparison methods include both traditional image processing methods (Histogram Equalization~\cite{DIP}, LIME~\cite{LIME}) and deep learning methods (DPED~\cite{DPED}, DPE~\cite{DPE_CVPR18}, RetinexNet~\cite{LOL_dataset}, Deep-UPE~\cite{DeepUPE_2019_CVPR}, Zero-DCE~\cite{zero_dce}, MSEC~\cite{Exposure_2021_CVPR}). Evaluation metrics are same as~\cite{Exposure_2021_CVPR}, including PSNR, SSIM and perceptual index (PI). Table.~\ref{tab:Exposure} shows that our \textbf{IAT} model has gained best result on all evaluation indexs. Compared to the second best result MSEC~\cite{Exposure_2021_CVPR},  IAT has much fewer parameters (0.09M v.s. 7M) and less evaluation time (0.004s per image v.s. 0.5s per image). Qualitative result has been shown in Fig.\ref{fig:low_level} and more visual results are given in supplementary material.   

\begin{table}[t]
\caption{Experimental results on low-light detection dataset EXDark~\cite{EXDark}, low-light semantic segmentation dataset ACDC~\cite{ACDC} and various light  detection dataset TYOL~\cite{Toyota_light}.}
\centering
\setlength\tabcolsep{2pt}
\begin{adjustbox}{max width = 0.95\linewidth}
\begin{tabular}{l|cc|cc|cc}
\Xhline{1.0pt}
\multirow{2}{*}{Methods} & \multicolumn{2}{c|}{$(d)$ EXDark Detection~\cite{EXDark}} & \multicolumn{2}{c|}{$(e)$ ACDC Segmentation~\cite{ACDC}} & \multicolumn{2}{c}{$(f)$ TYOL Detection~\cite{Toyota_light}} \\ \cline{2-7}  & \qquad mAP$\uparrow$ & time(s)$\downarrow$  & \qquad mIOU$\uparrow$  & time(s)$\downarrow$  & \qquad mAP$\uparrow$   & time(s)$\downarrow$ \\ \Xhline{0.6pt}
base-line  & \qquad 76.4  & 0.033  & \qquad \textbf{63.3}   & 0.249    & \qquad 88.4    & 0.023               \\ 
MBLLEN~\cite{Lv2018MBLLEN}     & \qquad 76.3    & 0.086     & \qquad 63.0   & 0.332  & \qquad 95.3             & 0.105               \\ 
DeepLPF~\cite{Deep_LPF}                  & \qquad 76.3              & 0.138                 & \qquad 61.9               & 0.807                 & \qquad 94.5             & 0.223               \\ 
Zero-DCE~\cite{zero_dce}                 & \qquad 76.9             & 0.042                 & \qquad 61.9               & 0.300                 & \qquad 95.2             & 0.030               \\ 
\textbf{IAT}                      & \qquad \textbf{77.2}              & \textbf{0.040}                 & \qquad 62.1              & \textbf{0.280 }                & \qquad \textbf{95.8}     & \textbf{0.027}       \\ \Xhline{1.0pt}
\end{tabular}
\end{adjustbox}
\label{tab:high}
\end{table}

\subsection{High-level Vision}
\label{exp:highlevel}

For high-level vision tasks: $\left\{ (d), (e), (f) \right\}$, we use IAT to restore the image before feeding to the subsequent recognition algorithms based on mmdetection and mmsegmentation ~\cite{mmdetection,mmseg2020}. For a fair comparison, we run all of the experiments in the same setting: same input size, same data augmentation methods (expand, random crop, multi-size, random flip...), same training epochs and same initial weights. We train the recognition algorithm on the datasets enhanced by IAT. We compare our methods with original datasets as well as datasets enhanced by other enhancement methods~\cite{Lv2018MBLLEN,Deep_LPF,zero_dce}. 

For object detection task in $(d)$ EXDark dataset~\cite{EXDark} and $(f)$ TYOL dataset~\cite{Toyota_light}. EXDark includes 7,363 real-world low-light images, ranging from twilight to extreme dark environment with 12 object categories. We take 
80$\%$ images of each category for training and the other 20$\%$ for testing.  TYOL includes 1680 images with 21 classes. We take 1365 images for training and other for evaluation. For both datasets, we perform object detection with YOLO-V3~\cite{yolov3}, all the input images have been cropped and resized to 608 $\times$ 608 pixel size, we use SGD optimizer to train YOLO-V3 with batch size 8 for 25 epochs to EXDark and 45 epochs to TYO-L, the initial learning rate is 1$e^{-3}$ and weight decay is 1$e^{-4}$. The detection metric mAP and per-image evaluation time is shown in Table.~\ref{tab:high}. Our IAT model gains best results in both accuracy and speed compared to the baseline model and other enhancement methods~\cite{Lv2018MBLLEN,Deep_LPF,zero_dce}.  

For semantic segmentation in $(e)$ ACDC dataset~\cite{ACDC}, we take 1006 night images in the ACDC dataset and then adopt DeepLab-V3+~\cite{Deeplabv3plus} to train on the ACDC-night train set and test on ACDC-night val set. The DeepLab-V3+~\cite{Deeplabv3plus} model is initialed by an Cityscape dataset~\cite{Cordts2016Cityscapes} pre-train model, we tuned the pre-train model by SGD optimizer with batch size 8 for 20000 iters, initial learning rate is set to 0.05, momentum and weight decay are separately set to 0.9 and 5$e^{-4}$. We show the segmentation metric mIOU and per-image evaluation time in Table.~\ref{tab:high}, we found that all the enhancements methods invalid in this setting, this may because the lightness condition in ACDC~\cite{ACDC} is various and exceeds the generalisation ability of the enhancement model. For this problem, we propose to joint training our IAT model with following segmentation network (as well as detection network), which would solves this problem and improve the semantic segmentation/ object detection results in low-light conditions, detailed analyse please refer to Sec.~\ref{sec:joint-train} of supplementary material~\footnote{For more experimental details and ablation analyse, please refer to the supplementary material.}.    
\vspace{-2mm}

\section{Conclusion}
We propose a novel lightweight transformer framework IAT, by adapting ISP-related parameters to adapt to  challenging light conditions. Despite its superior performance on several real-world datasets for both low-level and high-level tasks, IAT is extremely light with a fast speed. The lightweight and mobile-friendly IAT has the potential to become a standing tool for the computer vision community. 

However, one  mian drawback of the IAT module is that, the image signal processor (ISP) has been simplified due to the light-weight demand, we think that more detailed ISP-related parts could be concerned and interpolate to the IAT module. In further, we'd also like to implement IAT on 3D human relighting task, to solve more complex lighting problems under 3D condition. 
\vspace{-2mm}

\section{Acknowledgement}
Corner symbol '*' in the author name means the corresponding author. 
This work supported by JST Moonshot R$\&$D Grant
Number JPMJMS2011 and JST, ACT-X Grant Number
JPMJAX190D, Japan. This work also supported by National Natural Science Foundation of China (Grant No.  62206272) and Shanghai Committee of Science and Technology (Grant No. 21DZ1100100).

\bibliography{egbib}

\begin{thebibliography}{78}
\providecommand{\natexlab}[1]{#1}
\providecommand{\url}[1]{\texttt{#1}}
\expandafter\ifx\csname urlstyle\endcsname\relax
  \providecommand{\doi}[1]{doi: #1}\else
  \providecommand{\doi}{doi: \begingroup \urlstyle{rm}\Url}\fi

\bibitem[Afifi and Brown(2020)]{Afifi_2020_CVPR}
Mahmoud Afifi and Michael~S. Brown.
\newblock Deep white-balance editing.
\newblock In \emph{Proceedings of the IEEE/CVF Conference on Computer Vision
  and Pattern Recognition (CVPR)}, June 2020.

\bibitem[Afifi et~al.(2021)Afifi, Derpanis, Ommer, and
  Brown]{Exposure_2021_CVPR}
Mahmoud Afifi, Konstantinos~G. Derpanis, Bjorn Ommer, and Michael~S. Brown.
\newblock Learning multi-scale photo exposure correction.
\newblock In \emph{Proceedings of the IEEE/CVF Conference on Computer Vision
  and Pattern Recognition}, 2021.

\bibitem[Afifi et~al.(2022)Afifi, Brubaker, and Brown]{afifi2022awb}
Mahmoud Afifi, Marcus~A. Brubaker, and Michael~S. Brown.
\newblock Auto white-balance correction for mixed-illuminant scenes.
\newblock In \emph{IEEE Winter Conference on Applications of Computer Vision
  (WACV)}, 2022.

\bibitem[Ba et~al.(2016)Ba, Kiros, and Hinton]{ln}
Jimmy Ba, Jamie~Ryan Kiros, and Geoffrey~E. Hinton.
\newblock Layer normalization.
\newblock \emph{ArXiv}, abs/1607.06450, 2016.

\bibitem[Brooks et~al.(2019)Brooks, Mildenhall, Xue, Chen, Sharlet, and
  Barron]{brooks2019unprocessing}
Tim Brooks, Ben Mildenhall, Tianfan Xue, Jiawen Chen, Dillon Sharlet, and
  Jonathan~T Barron.
\newblock Unprocessing images for learned raw denoising.
\newblock In \emph{Proceedings of the IEEE Conference on Computer Vision and
  Pattern Recognition}, 2019.

\bibitem[Bychkovsky et~al.(2011)Bychkovsky, Paris, Chan, and
  Durand]{fivek_dataset}
Vladimir Bychkovsky, Sylvain Paris, Eric Chan, and Fr{\'e}do Durand.
\newblock Learning photographic global tonal adjustment with a database of
  input / output image pairs.
\newblock In \emph{Proceedings of the IEEE Conference on Computer Vision and
  Pattern Recognition}, 2011.

\bibitem[Carion et~al.(2020)Carion, Massa, Synnaeve, Usunier, Kirillov, and
  Zagoruyko]{DETR}
Nicolas Carion, Francisco Massa, Gabriel Synnaeve, Nicolas Usunier, Alexander
  Kirillov, and Sergey Zagoruyko.
\newblock End-to-end object detection with transformers.
\newblock In \emph{European conference on computer vision}, 2020.

\bibitem[{Chen} et~al.(2018){Chen}, {Chen}, {Xu}, and
  {Koltun}]{see_in_the_dark}
C.~{Chen}, Q.~{Chen}, J.~{Xu}, and V.~{Koltun}.
\newblock Learning to see in the dark.
\newblock In \emph{IEEE/CVF Conference on Computer Vision and Pattern
  Recognition}, 2018.

\bibitem[Chen et~al.(2021{\natexlab{a}})Chen, Wang, Guo, Xu, Deng, Liu, Ma, Xu,
  Xu, and Gao]{IPT_CVPR}
Hanting Chen, Yunhe Wang, Tianyu Guo, Chang Xu, Yiping Deng, Zhenhua Liu, Siwei
  Ma, Chunjing Xu, Chao Xu, and Wen Gao.
\newblock Pre-trained image processing transformer.
\newblock In \emph{Proceedings of the IEEE/CVF Conference on Computer Vision
  and Pattern Recognition}, 2021{\natexlab{a}}.

\bibitem[Chen et~al.(2019)Chen, Wang, Pang, Cao, Xiong, Li, Sun, Feng, Liu, Xu,
  Zhang, Cheng, Zhu, Cheng, Zhao, Li, Lu, Zhu, Wu, Dai, Wang, Shi, Ouyang, Loy,
  and Lin]{mmdetection}
Kai Chen, Jiaqi Wang, Jiangmiao Pang, Yuhang Cao, Yu~Xiong, Xiaoxiao Li,
  Shuyang Sun, Wansen Feng, Ziwei Liu, Jiarui Xu, Zheng Zhang, Dazhi Cheng,
  Chenchen Zhu, Tianheng Cheng, Qijie Zhao, Buyu Li, Xin Lu, Rui Zhu, Yue Wu,
  Jifeng Dai, Jingdong Wang, Jianping Shi, Wanli Ouyang, Chen~Change Loy, and
  Dahua Lin.
\newblock {MMDetection}: Open mmlab detection toolbox and benchmark.
\newblock \emph{arXiv preprint arXiv:1906.07155}, 2019.

\bibitem[Chen et~al.(2018{\natexlab{a}})Chen, Zhu, Papandreou, Schroff, and
  Adam]{Deeplabv3plus}
Liang-Chieh Chen, Yukun Zhu, George Papandreou, Florian Schroff, and Hartwig
  Adam.
\newblock Encoder-decoder with atrous separable convolution for semantic image
  segmentation.
\newblock In \emph{Proceedings of the European conference on computer vision},
  2018{\natexlab{a}}.

\bibitem[Chen et~al.(2021{\natexlab{b}})Chen, Feng, Gao, Xu, and
  Chen]{ISP_uprocess_2021_ICCV}
Shiqi Chen, Huajun Feng, Keming Gao, Zhihai Xu, and Yueting Chen.
\newblock Extreme-quality computational imaging via degradation framework.
\newblock In \emph{Proceedings of the IEEE/CVF International Conference on
  Computer Vision}, 2021{\natexlab{b}}.

\bibitem[Chen et~al.(2018{\natexlab{b}})Chen, Wang, Kao, and
  Chuang]{DPE_CVPR18}
Yu-Sheng Chen, Yu-Ching Wang, Man-Hsin Kao, and Yung-Yu Chuang.
\newblock Deep photo enhancer: Unpaired learning for image enhancement from
  photographs with gans.
\newblock In \emph{IEEE/CVF Conference on Computer Vision and Pattern
  Recognition}, 2018{\natexlab{b}}.

\bibitem[Contributors(2020)]{mmseg2020}
MMSegmentation Contributors.
\newblock {MMSegmentation}: Openmmlab semantic segmentation toolbox and
  benchmark.
\newblock \url{https://github.com/open-mmlab/mmsegmentation}, 2020.

\bibitem[Cordts et~al.(2016)Cordts, Omran, Ramos, Rehfeld, Enzweiler, Benenson,
  Franke, Roth, and Schiele]{Cordts2016Cityscapes}
Marius Cordts, Mohamed Omran, Sebastian Ramos, Timo Rehfeld, Markus Enzweiler,
  Rodrigo Benenson, Uwe Franke, Stefan Roth, and Bernt Schiele.
\newblock The cityscapes dataset for semantic urban scene understanding.
\newblock In \emph{Proc. of the IEEE Conference on Computer Vision and Pattern
  Recognition (CVPR)}, 2016.

\bibitem[Cui et~al.(2021)Cui, Qi, Gu, You, Zhang, and Harada]{ICCV_MAET}
Ziteng Cui, Guo-Jun Qi, Lin Gu, Shaodi You, Zenghui Zhang, and Tatsuya Harada.
\newblock Multitask aet with orthogonal tangent regularity for dark object
  detection.
\newblock In \emph{Proceedings of the IEEE/CVF International Conference on
  Computer Vision}, 2021.

\bibitem[Deng et~al.(2009)Deng, Dong, Socher, Li, Li, and Fei-Fei]{imagenet}
Jia Deng, Wei Dong, Richard Socher, Li-Jia Li, Kai Li, and Li~Fei-Fei.
\newblock Imagenet: A large-scale hierarchical image database.
\newblock In \emph{IEEE conference on computer vision and pattern recognition},
  2009.

\bibitem[Dosovitskiy et~al.(2021)Dosovitskiy, Beyer, Kolesnikov, Weissenborn,
  Zhai, Unterthiner, Dehghani, Minderer, Heigold, Gelly, Uszkoreit, and
  Houlsby]{ViT}
Alexey Dosovitskiy, Lucas Beyer, Alexander Kolesnikov, Dirk Weissenborn,
  Xiaohua Zhai, Thomas Unterthiner, Mostafa Dehghani, Matthias Minderer, Georg
  Heigold, Sylvain Gelly, Jakob Uszkoreit, and Neil Houlsby.
\newblock An image is worth 16x16 words: Transformers for image recognition at
  scale.
\newblock In \emph{International Conference on Learning Representations}, 2021.

\bibitem[Fu et~al.(2022)Fu, Hong, Chen, and You]{fu2022gan}
Ying Fu, Yang Hong, Linwei Chen, and Shaodi You.
\newblock Le-gan: Unsupervised low-light image enhancement network using
  attention module and identity invariant loss.
\newblock \emph{Knowledge-Based Systems}, 2022.

\bibitem[Gonzalez and Woods(2006)]{DIP}
Rafael~C. Gonzalez and Richard~E. Woods.
\newblock \emph{Digital Image Processing (3rd Edition)}.
\newblock Prentice-Hall, Inc., USA, 2006.
\newblock ISBN 013168728X.

\bibitem[{Guo} et~al.(2020){Guo}, {Li}, {Guo}, {Loy}, {Hou}, {Kwong}, and
  {Cong}]{zero_dce}
C.~{Guo}, C.~{Li}, J.~{Guo}, C.~C. {Loy}, J.~{Hou}, S.~{Kwong}, and R.~{Cong}.
\newblock Zero-reference deep curve estimation for low-light image enhancement.
\newblock In \emph{IEEE/CVF Conference on Computer Vision and Pattern
  Recognition}, 2020.

\bibitem[Guo et~al.(2017)Guo, Li, and Ling]{LIME}
Xiaojie Guo, Yu~Li, and Haibin Ling.
\newblock Lime: Low-light image enhancement via illumination map estimation.
\newblock \emph{IEEE Transactions on Image Processing}, 2017.

\bibitem[Han et~al.(2022)Han, Fan, Dai, Sun, Cheng, Liu, and
  Wang]{han2021connection}
Qi~Han, Zejia Fan, Qi~Dai, Lei Sun, Ming-Ming Cheng, Jiaying Liu, and Jingdong
  Wang.
\newblock On the connection between local attention and dynamic depth-wise
  convolution.
\newblock In \emph{International Conference on Learning Representations}, 2022.

\bibitem[He et~al.(2016)He, Zhang, Ren, and Sun]{ResNet}
Kaiming He, Xiangyu Zhang, Shaoqing Ren, and Jian Sun.
\newblock Deep residual learning for image recognition.
\newblock In \emph{2016 IEEE Conference on Computer Vision and Pattern
  Recognition}, 2016.

\bibitem[Heide and et.al(2014)]{ISP_flexisp}
Felix Heide and Steinberger et.al.
\newblock Flexisp: A flexible camera image processing framework.
\newblock \emph{ACM Trans. Graph.}, 2014.

\bibitem[Hodan et~al.(2018)Hodan, Michel, Brachmann, Kehl, GlentBuch, Kraft,
  Drost, Vidal, Ihrke, Zabulis, et~al.]{Toyota_light}
Tomas Hodan, Frank Michel, Eric Brachmann, Wadim Kehl, Anders GlentBuch, Dirk
  Kraft, Bertram Drost, Joel Vidal, Stephan Ihrke, Xenophon Zabulis, et~al.
\newblock Bop: Benchmark for 6d object pose estimation.
\newblock In \emph{Proceedings of the European Conference on Computer Vision},
  2018.

\bibitem[Hong et~al.(2021)Hong, Wei, Chen, and Fu]{Hong_2021_BMVC}
Yang Hong, Kaixuan Wei, Linwei Chen, and Ying Fu.
\newblock Crafting object detection in very low light.
\newblock In \emph{The British Machine Vision Conference}, November 2021.

\bibitem[Hu et~al.(2018)Hu, He, Xu, Wang, and Lin]{white_box}
Yuanming Hu, Hao He, Chenxi Xu, Baoyuan Wang, and Stephen Lin.
\newblock Exposure: A white-box photo post-processing framework.
\newblock \emph{ACM Trans. Graph.}, 2018.

\bibitem[Ignatov et~al.(2017)Ignatov, Kobyshev, Timofte, Vanhoey, and
  Van~Gool]{DPED}
Andrey Ignatov, Nikolay Kobyshev, Radu Timofte, Kenneth Vanhoey, and Luc
  Van~Gool.
\newblock Dslr-quality photos on mobile devices with deep convolutional
  networks.
\newblock In \emph{Proceedings of the IEEE international conference on computer
  vision}, 2017.

\bibitem[{Jiang} et~al.(2017){Jiang}, {Tian}, {Farrell}, and
  {Wandell}]{ISP_L3method}
H.~{Jiang}, Q.~{Tian}, J.~{Farrell}, and B.~A. {Wandell}.
\newblock Learning the image processing pipeline.
\newblock \emph{IEEE Transactions on Image Processing}, 2017.

\bibitem[Johnson et~al.(2016)Johnson, Alahi, and Fei-Fei]{Perceptual_loss}
Justin Johnson, Alexandre Alahi, and Li~Fei-Fei.
\newblock Perceptual losses for real-time style transfer and super-resolution.
\newblock In \emph{European Conference on Computer Vision}, 2016.

\bibitem[Karaimer and Brown(2016)]{ISP_pipeline_eccv16}
Hakki~Can Karaimer and Michael~S. Brown.
\newblock A software platform for manipulating the camera imaging pipeline.
\newblock In \emph{European Conference on Computer Vision}, 2016.

\bibitem[Kim et~al.(2021)Kim, Choi, Kim, and Koh]{RCT_ICCV21}
Hanul Kim, Su-Min Choi, Chang-Su Kim, and Yeong~Jun Koh.
\newblock Representative color transform for image enhancement.
\newblock In \emph{Proceedings of the IEEE/CVF International Conference on
  Computer Vision}, 2021.

\bibitem[Kumar et~al.(2021)Kumar, Weissenborn, and
  Kalchbrenner]{kumar2021colorization}
Manoj Kumar, Dirk Weissenborn, and Nal Kalchbrenner.
\newblock Colorization transformer.
\newblock In \emph{International Conference on Learning Representations}, 2021.

\bibitem[Land(1986)]{retinex}
Edwin~H. Land.
\newblock An alternative technique for the computation of the designator in the
  retinex theory of color vision.
\newblock \emph{Proceedings of the National Academy of Sciences of the United
  States of America}, 1986.

\bibitem[Li et~al.(2022)Li, Wang, Zhang, Gao, Song, Liu, Li, and
  Qiao]{uniformer1}
Kunchang Li, Yali Wang, Junhao Zhang, Peng Gao, Guanglu Song, Yu~Liu, Hongsheng
  Li, and Yu~Qiao.
\newblock Uniformer: Unifying convolution and self-attention for visual
  recognition.
\newblock \emph{arXiv preprint arXiv:2201.09450}, 2022.

\bibitem[Liang et~al.(2021)Liang, Cao, Sun, Zhang, Van~Gool, and
  Timofte]{liang2021swinir}
Jingyun Liang, Jiezhang Cao, Guolei Sun, Kai Zhang, Luc Van~Gool, and Radu
  Timofte.
\newblock Swinir: Image restoration using swin transformer.
\newblock In \emph{IEEE International Conference on Computer Vision Workshops},
  2021.

\bibitem[Lin et~al.(2014)Lin, Maire, Belongie, Hays, Perona, Ramanan,
  Doll{\'a}r, and Zitnick]{coco_dataset}
Tsung-Yi Lin, Michael Maire, Serge Belongie, James Hays, Pietro Perona, Deva
  Ramanan, Piotr Doll{\'a}r, and C~Lawrence Zitnick.
\newblock Microsoft coco: Common objects in context.
\newblock In \emph{European conference on computer vision}, 2014.

\bibitem[Liu et~al.(2021{\natexlab{a}})Liu, Xu, Yang, Fan, and
  Huang]{IJCV_21_lowface}
Jiaying Liu, Dejia Xu, Wenhan Yang, Minhao Fan, and Haofeng Huang.
\newblock Benchmarking low-light image enhancement and beyond.
\newblock \emph{International Journal of Computer Vision}, 2021{\natexlab{a}}.

\bibitem[Liu et~al.(2021{\natexlab{b}})Liu, Lin, Cao, Hu, Wei, Zhang, Lin, and
  Guo]{swin}
Ze~Liu, Yutong Lin, Yue Cao, Han Hu, Yixuan Wei, Zheng Zhang, Stephen Lin, and
  Baining Guo.
\newblock Swin transformer: Hierarchical vision transformer using shifted
  windows.
\newblock In \emph{Proceedings of the IEEE/CVF International Conference on
  Computer Vision}, 2021{\natexlab{b}}.

\bibitem[Loh and Chan(2019)]{EXDark}
Yuen~Peng Loh and Chee~Seng Chan.
\newblock Getting to know low-light images with the exclusively dark dataset.
\newblock \emph{Computer Vision and Image Understanding}, 2019.

\bibitem[Lore et~al.(2017)Lore, Akintayo, and Sarkar]{LLNet}
Kin~Gwn Lore, Adedotun Akintayo, and Soumik Sarkar.
\newblock Llnet: A deep autoencoder approach to natural low-light image
  enhancement.
\newblock \emph{Pattern Recognition}, 2017.

\bibitem[Lv et~al.(2018)Lv, Lu, Wu, and Lim]{Lv2018MBLLEN}
Feifan Lv, Feng Lu, Jianhua Wu, and Chongsoon Lim.
\newblock Mbllen: Low-light image/video enhancement using cnns.
\newblock In \emph{British Machine Vision Conference}, 2018.

\bibitem[Minciullo et~al.(2021)Minciullo, Manhardt, Yoshikawa, Meier, Tombari,
  and Kobori]{DBGAN_2021_WACV_strong_light}
Luca Minciullo, Fabian Manhardt, Kei Yoshikawa, Sven Meier, Federico Tombari,
  and Norimasa Kobori.
\newblock Db-gan: Boosting object recognition under strong lighting conditions.
\newblock In \emph{Proceedings of the IEEE/CVF Winter Conference on
  Applications of Computer Vision}, 2021.

\bibitem[Moran et~al.(2020)Moran, Marza, McDonagh, Parisot, and
  Slabaugh]{Deep_LPF}
Sean Moran, Pierre Marza, Steven McDonagh, Sarah Parisot, and Gregory Slabaugh.
\newblock Deeplpf: Deep local parametric filters for image enhancement.
\newblock In \emph{Proceedings of the IEEE/CVF Conference on Computer Vision
  and Pattern Recognition}, 2020.

\bibitem[Nayar and Branzoi(2003)]{exposure_ICCV2003}
Nayar and Branzoi.
\newblock Adaptive dynamic range imaging: optical control of pixel exposures
  over space and time.
\newblock In \emph{Proceedings Ninth IEEE International Conference on Computer
  Vision}, pages 1168--1175 vol.2, 2003.
\newblock \doi{10.1109/ICCV.2003.1238624}.

\bibitem[Nsampi et~al.(2021)Nsampi, Hu, and Wang]{exposure_bmvc21}
Ntumba~Elie Nsampi, Zhongyun Hu, and Qing Wang.
\newblock Learning exposure correction via consistency modeling.
\newblock In \emph{BMVC}, 2021.

\bibitem[Park et~al.(2018)Park, Lee, Yoo, and Kweon]{park2018distort}
Jongchan Park, Joon-Young Lee, Donggeun Yoo, and In~So Kweon.
\newblock Distort-and-recover: Color enhancement using deep reinforcement
  learning.
\newblock In \emph{Proceedings of the IEEE conference on computer vision and
  pattern recognition}, 2018.

\bibitem[Radford et~al.(2021)Radford, Kim, Hallacy, Ramesh, Goh, Agarwal,
  Sastry, Askell, Mishkin, Clark, Krueger, and Sutskever]{clip}
Alec Radford, Jong~Wook Kim, Chris Hallacy, Aditya Ramesh, Gabriel Goh,
  Sandhini Agarwal, Girish Sastry, Amanda Askell, Pamela Mishkin, Jack Clark,
  Gretchen Krueger, and Ilya Sutskever.
\newblock Learning transferable visual models from natural language
  supervision.
\newblock In Marina Meila and Tong Zhang, editors, \emph{Proceedings of the
  38th International Conference on Machine Learning}, volume 139 of
  \emph{Proceedings of Machine Learning Research}, pages 8748--8763. PMLR,
  18--24 Jul 2021.

\bibitem[Redmon and Farhadi(2018)]{yolov3}
Joseph Redmon and Ali Farhadi.
\newblock Yolov3: An incremental improvement.
\newblock \emph{arXiv preprint arXiv:1804.02767}, 2018.

\bibitem[Ronneberger et~al.(2015)Ronneberger, Fischer, and Brox]{unet}
Olaf Ronneberger, Philipp Fischer, and Thomas Brox.
\newblock U-net: Convolutional networks for biomedical image segmentation,
  2015.

\bibitem[Sakaridis et~al.(2021)Sakaridis, Dai, and Van~Gool]{ACDC}
Christos Sakaridis, Dengxin Dai, and Luc Van~Gool.
\newblock {ACDC}: The adverse conditions dataset with correspondences for
  semantic driving scene understanding.
\newblock In \emph{Proceedings of the IEEE/CVF International Conference on
  Computer Vision}, 2021.

\bibitem[Sasagawa and Nagahara(2020)]{yolo-in-the-dark}
Yukihiro Sasagawa and Hajime Nagahara.
\newblock Yolo in the dark: Domain adaptation method for merging multiple
  models.
\newblock In \emph{Proceedings of European Conference on Computer Vision},
  2020.

\bibitem[Sharma and Tan(2021)]{Tan_enhancement}
Aashish Sharma and Robby~T. Tan.
\newblock Nighttime visibility enhancement by increasing the dynamic range and
  suppression of light effects.
\newblock In \emph{2021 IEEE/CVF Conference on Computer Vision and Pattern
  Recognition (CVPR)}, pages 11972--11981, 2021.
\newblock \doi{10.1109/CVPR46437.2021.01180}.

\bibitem[Tolstikhin et~al.(2021)Tolstikhin, Houlsby, Kolesnikov, Beyer, Zhai,
  Unterthiner, Yung, Steiner, Keysers, Uszkoreit, Lucic, and
  Dosovitskiy]{mlpmixer}
Ilya~O Tolstikhin, Neil Houlsby, Alexander Kolesnikov, Lucas Beyer, Xiaohua
  Zhai, Thomas Unterthiner, Jessica Yung, Andreas Steiner, Daniel Keysers,
  Jakob Uszkoreit, Mario Lucic, and Alexey Dosovitskiy.
\newblock Mlp-mixer: An all-mlp architecture for vision.
\newblock In M.~Ranzato, A.~Beygelzimer, Y.~Dauphin, P.S. Liang, and J.~Wortman
  Vaughan, editors, \emph{Advances in Neural Information Processing Systems},
  volume~34, pages 24261--24272. Curran Associates, Inc., 2021.
\newblock URL
  \url{https://proceedings.neurips.cc/paper/2021/file/cba0a4ee5ccd02fda0fe3f9a3e7b89fe-Paper.pdf}.

\bibitem[Tomasi and Manduchi(1998)]{Global_HE}
Carlo Tomasi and Roberto Manduchi.
\newblock Bilateral filtering for gray and color images.
\newblock In \emph{Sixth international conference on computer vision (IEEE Cat.
  No. 98CH36271)}, 1998.

\bibitem[Touvron et~al.(2021{\natexlab{a}})Touvron, Bojanowski, Caron, Cord,
  El-Nouby, Grave, Izacard, Joulin, Synnaeve, Verbeek,
  et~al.]{touvron2021resmlp}
Hugo Touvron, Piotr Bojanowski, Mathilde Caron, Matthieu Cord, Alaaeldin
  El-Nouby, Edouard Grave, Gautier Izacard, Armand Joulin, Gabriel Synnaeve,
  Jakob Verbeek, et~al.
\newblock Resmlp: Feedforward networks for image classification with
  data-efficient training.
\newblock \emph{arXiv preprint arXiv:2105.03404}, 2021{\natexlab{a}}.

\bibitem[Touvron et~al.(2021{\natexlab{b}})Touvron, Cord, Sablayrolles,
  Synnaeve, and J{\'e}gou]{cait}
Hugo Touvron, Matthieu Cord, Alexandre Sablayrolles, Gabriel Synnaeve, and
  Herv{\'e} J{\'e}gou.
\newblock Going deeper with image transformers.
\newblock In \emph{Proceedings of the IEEE/CVF International Conference on
  Computer Vision}, 2021{\natexlab{b}}.

\bibitem[Tu et~al.(2022)Tu, Talebi, Zhang, Yang, Milanfar, Bovik, and
  Li]{MLP_enhancement}
Zhengzhong Tu, Hossein Talebi, Han Zhang, Feng Yang, Peyman Milanfar, Alan
  Bovik, and Yinxiao Li.
\newblock Maxim: Multi-axis mlp for image processing.
\newblock \emph{CVPR}, 2022.

\bibitem[Valanarasu et~al.(2021)Valanarasu, Yasarla, and Patel]{TransWeather}
Jeya Maria~Jose Valanarasu, Rajeev Yasarla, and Vishal~M Patel.
\newblock Transweather: Transformer-based restoration of images degraded by
  adverse weather conditions.
\newblock \emph{arXiv preprint arXiv:2111.14813}, 2021.

\bibitem[Vaswani et~al.(2017)Vaswani, Shazeer, Parmar, Uszkoreit, Jones, Gomez,
  Kaiser, and Polosukhin]{Transformer_NLP}
Ashish Vaswani, Noam Shazeer, Niki Parmar, Jakob Uszkoreit, Llion Jones,
  Aidan~N Gomez, \L~ukasz Kaiser, and Illia Polosukhin.
\newblock Attention is all you need.
\newblock In I.~Guyon, U.~Von Luxburg, S.~Bengio, H.~Wallach, R.~Fergus,
  S.~Vishwanathan, and R.~Garnett, editors, \emph{Advances in Neural
  Information Processing Systems}, volume~30. Curran Associates, Inc., 2017.
\newblock URL
  \url{https://proceedings.neurips.cc/paper/2017/file/3f5ee243547dee91fbd053c1c4a845aa-Paper.pdf}.

\bibitem[Wang et~al.(2019)Wang, Zhang, Fu, Shen, Zheng, and
  Jia]{DeepUPE_2019_CVPR}
Ruixing Wang, Qing Zhang, Chi-Wing Fu, Xiaoyong Shen, Wei-Shi Zheng, and Jiaya
  Jia.
\newblock Underexposed photo enhancement using deep illumination estimation.
\newblock In \emph{The IEEE Conference on Computer Vision and Pattern
  Recognition}, 2019.

\bibitem[Wang et~al.(2022)Wang, Cun, Bao, Zhou, Liu, and Li]{Wang2022Uformer}
Zhendong Wang, Xiaodong Cun, Jianmin Bao, Wengang Zhou, Jianzhuang Liu, and
  Houqiang Li.
\newblock Uformer: A general u-shaped transformer for image restoration.
\newblock In \emph{Proceedings of the IEEE Conference on Computer Vision and
  Pattern Recognition (CVPR)}, 2022.

\bibitem[Wei et~al.(2018)Wei, Wang, Yang, and Liu]{LOL_dataset}
Chen Wei, Wenjing Wang, Wenhan Yang, and Jiaying Liu.
\newblock Deep retinex decomposition for low-light enhancement.
\newblock In \emph{British Machine Vision Conference}, 2018.

\bibitem[Wei et~al.(2020)Wei, Fu, Yang, and
  Huang]{CVPRLow_light_denoise_physics}
Kaixuan Wei, Ying Fu, Jiaolong Yang, and Hua Huang.
\newblock A physics-based noise formation model for extreme low-light raw
  denoising.
\newblock In \emph{IEEE Conference on Computer Vision and Pattern Recognition},
  2020.

\bibitem[Xie et~al.(2021)Xie, Wang, Yu, Anandkumar, Alvarez, and
  Luo]{xie2021segformer}
Enze Xie, Wenhai Wang, Zhiding Yu, Anima Anandkumar, Jose~M Alvarez, and Ping
  Luo.
\newblock Segformer: Simple and efficient design for semantic segmentation with
  transformers.
\newblock \emph{arXiv preprint arXiv:2105.15203}, 2021.

\bibitem[Xing et~al.(2021)Xing, Qian, and Chen]{xing21invertible}
Yazhou Xing, Zian Qian, and Qifeng Chen.
\newblock Invertible image signal processing.
\newblock In \emph{CVPR}, 2021.

\bibitem[Xu et~al.(2022)Xu, Wang, Fu, and Jia]{SNR_2022_CVPR}
Xiaogang Xu, Ruixing Wang, Chi-Wing Fu, and Jiaya Jia.
\newblock Snr-aware low-light image enhancement.
\newblock In \emph{Proceedings of the IEEE/CVF Conference on Computer Vision
  and Pattern Recognition (CVPR)}, pages 17714--17724, June 2022.

\bibitem[Yang et~al.(2022)Yang, Cheng, Zhao, Zhang, and Li]{Learn_adapt_light}
Kai-Fu Yang, Cheng Cheng, Shi-Xuan Zhao, Xian-Shi Zhang, and Yong-Jie Li.
\newblock Learning to adapt to light, 2022.
\newblock URL \url{https://arxiv.org/abs/2202.08098}.

\bibitem[Yang et~al.(2020)Yang, Wang, Fang, Wang, and Liu]{DRBN_CVPR}
Wenhan Yang, Shiqi Wang, Yuming Fang, Yue Wang, and Jiaying Liu.
\newblock From fidelity to perceptual quality: A semi-supervised approach for
  low-light image enhancement.
\newblock In \emph{IEEE/CVF Conference on Computer Vision and Pattern
  Recognition}, 2020.

\bibitem[Yu et~al.(2018)Yu, Liu, Zhang, Qu, Zhao, and Zhang]{YuNIPS18}
Runsheng Yu, Wenyu Liu, Yasen Zhang, Zhi Qu, Deli Zhao, and Bo~Zhang.
\newblock Deepexposure: Learning to expose photos with asynchronously
  reinforced adversarial learning.
\newblock In \emph{Advances in Neural Information Processing Systems}, 2018.

\bibitem[Yuan and Sun(2012)]{Exposure_ECCV12}
Lu~Yuan and Jian Sun.
\newblock Automatic exposure correction of consumer photographs.
\newblock In \emph{European Conference on Computer Vision}, 2012.

\bibitem[Zamir et~al.(2021)Zamir, Arora, Khan, Hayat, Khan, and
  Yang]{zamir2021restormer}
Syed~Waqas Zamir, Aditya Arora, Salman Khan, Munawar Hayat, Fahad~Shahbaz Khan,
  and Ming-Hsuan Yang.
\newblock Restormer: Efficient transformer for high-resolution image
  restoration.
\newblock \emph{arXiv preprint arXiv:2111.09881}, 2021.

\bibitem[Zeng et~al.(2020)Zeng, Cai, Li, Cao, and Zhang]{3DLUT}
Hui Zeng, Jianrui Cai, Lida Li, Zisheng Cao, and Lei Zhang.
\newblock Learning image-adaptive 3d lookup tables for high performance photo
  enhancement in real-time.
\newblock \emph{IEEE Transactions on Pattern Analysis and Machine
  Intelligence}, pages 1--1, 2020.
\newblock \doi{10.1109/TPAMI.2020.3026740}.

\bibitem[Zhang et~al.(2021)Zhang, Fang, Gao, Zhang, Li, Dai, Qiao, and
  Li]{zhang2021tip}
Renrui Zhang, Rongyao Fang, Peng Gao, Wei Zhang, Kunchang Li, Jifeng Dai,
  Yu~Qiao, and Hongsheng Li.
\newblock Tip-adapter: Training-free clip-adapter for better vision-language
  modeling.
\newblock \emph{arXiv preprint arXiv:2111.03930}, 2021.

\bibitem[Zhang et~al.(2022)Zhang, Qiu, Wang, Xu, Guo, Qiao, Gao, and
  Li]{zhang2022monodetr}
Renrui Zhang, Han Qiu, Tai Wang, Xuanzhuo Xu, Ziyu Guo, Yu~Qiao, Peng Gao, and
  Hongsheng Li.
\newblock Monodetr: Depth-aware transformer for monocular 3d object detection.
\newblock \emph{arXiv preprint arXiv:2203.13310}, 2022.

\bibitem[Zhang et~al.(2019)Zhang, Zhang, and Guo]{KIND}
Yonghua Zhang, Jiawan Zhang, and Xiaojie Guo.
\newblock Kindling the darkness: A practical low-light image enhancer.
\newblock In \emph{Proceedings of the 27th ACM international conference on
  multimedia}, 2019.

\bibitem[Zheng et~al.(2020)Zheng, Zhang, and Lu]{Optical_Flow_in_dark}
Yinqiang Zheng, Mingfang Zhang, and Feng Lu.
\newblock Optical flow in the dark.
\newblock In \emph{Proceedings of the IEEE/CVF Conference on Computer Vision
  and Pattern Recognition}, pages 6749--6757, 2020.

\end{thebibliography}
\clearpage

\section{Analyse on Module Structure}

For the global part $g$ of the IAT module, here we simplify the ISP procedures~\cite{brooks2019unprocessing,ICCV_MAET,ISP_uprocess_2021_ICCV} as the following equation:

\begin{equation}
G(\cdot) = Gamma(W_{ccm} ( W_{wb} (\cdot))).
\label{eq:ISP_Function_detail_supp}
\end{equation}

White balance (WB) function is an essential part in ISP pipeline. WB algorithm estimates the per channel gain on the image, to maintain the object's colour constancy under various different light colour. WB is usually represented as a $3 \times 3$ diagonal von Kris matrix $W_{wb}$ in camera imaging pipeline~\cite{Afifi_2020_CVPR,afifi2022awb,brooks2019unprocessing,ISP_pipeline_eccv16}. After that, camera color matrix (CCM) $W_{ccm}$ converts the white-balanced data from camera internal color space cRGB to sRGB colour space~\cite{ISP_pipeline_eccv16,ICCV_MAET,brooks2019unprocessing}. At last gamma correction aims to match non-linearity of humans perception on dark regions. A standard gamma curve is usually represent as an exponential function with the exponential parameter $\gamma$, so we build our global branch $g_t(\cdot)$ following the equation:

\begin{equation}
g_t(\cdot) = (max(\sum_{c_j} W_{c_i,c_j} (\cdot), \epsilon))^\gamma,  c_i,c_j \in \{r,g,b\},
\label{eq:ISP_Function_global}
\end{equation}
where the $W_{c_i,c_j}$ is a joint colour transform function consist of white balance matrix and CCM and $\gamma$ is the gamma correction's exponential value, $\epsilon$ is a minimum number to keep non-negative. Final as we discussed in Sec.3.1, the input image $I_i$ would separately pass by local branch $f$ and global branch $g$ to generate the prediction result $\hat{I_t} = g_t(f(I_i))$.

\vspace{1mm}

We also evaluate to train the model with corresponding raw-RGB data as additional supervision. Since it's hard to directly get raw-RGB data from the currect dataset, we then adopt the Invertible ISP~\cite{xing21invertible} to generate corresponding raw-RGB data $I_{raw}$ from the input image $I_i$, we use pre-train weights in~\cite{xing21invertible}  to generate $I_{raw}$. In the training stage, we additional add a loss function $L_{raw}$ for raw-RGB supervision, the total loss function shown as follow:

\begin{equation}
\begin{aligned}
L_{total} = & L_{rgb} +  \lambda \cdot L_{raw}\\
= & L_1(g_t(f(I_i), I_t) + \lambda \cdot L_1(f(I_i), I_{raw}).
\label{eq:Loss}
\end{aligned}
\end{equation}

$L_{total}$ is the total loss function that consist of two parts: the first part $L_{rgb}$ is L1 loss function between predict result $g_t(f(I_i))$ with ground truth image $I_t$, while the second part $L_{raw}$ is the L1 loss function between middle representation $f(I_i)$ and raw-RGB image $I_{raw}$ for raw-RGB part supervision, and $\lambda$ is a balance parameter where we set it to 0.1 in our experiments. We make the comparison experiments on exposure correction dataset~\cite{Exposure_2021_CVPR}, the training and experiments’ settings are follow the settings in Sec.4.2, only difference is the training strategy with or without raw-RGB supervision. The comparison results are shown in Table~\ref{Tab:RAW-RGB}, we can find that with the additional supervision of raw-RGB data, most of evaluation metrics on exposure correction dataset~\cite{Exposure_2021_CVPR} would be improved.

\begin{table}[t]
\caption{Comparison experiments of with (w) and without (w/o) raw-RGB supervision on exposure correction dataset~\cite{Exposure_2021_CVPR}.}
\centering
\begin{adjustbox}{max width = 0.95\linewidth}
\begin{tabular}{c|cc|cc|cc|cc|cc}
\Xhline{1.0pt}
\multirow{2}{*}{Method} & \multicolumn{2}{c|}{Expert A} & \multicolumn{2}{c|}{Expert B} & \multicolumn{2}{c|}{Expert C} & \multicolumn{2}{c|}{Expert D} & \multicolumn{2}{c}{Expert E} \\ \cline{2-11} 
                        & PSNR$\uparrow$          & SSIM$\uparrow$        & PSNR$\uparrow$          & SSIM$\uparrow$           & PSNR$\uparrow$           & SSIM$\uparrow$           & PSNR$\uparrow$           & SSIM$\uparrow$           & PSNR$\uparrow$           & SSIM$\uparrow$          \\ \Xhline{0.6pt}
w/o raw-RGB             & 19.90         & 0.817         & 21.65         & 0.867         & \textbf{21.23}         & \textbf{0.850}        & 19.86         & 0.844         & 19.34         & 0.840        \\ \hline
w raw-RGB               & \textbf{19.98}   & \textbf{0.822}     & \textbf{22.03}        & \textbf{0.885}         & 21.16         & 0.843         & \textbf{19.94}        & \textbf{0.852}         & \textbf{19.48}         & \textbf{0.841}        \\ \Xhline{1.0pt}
\end{tabular}
\end{adjustbox}
\label{Tab:RAW-RGB}
\end{table}

\section{Joint Training with High-level Framework}
\label{sec:joint-train}
\begin{figure}[h]
    \centering
    \includegraphics[width=0.9\linewidth]{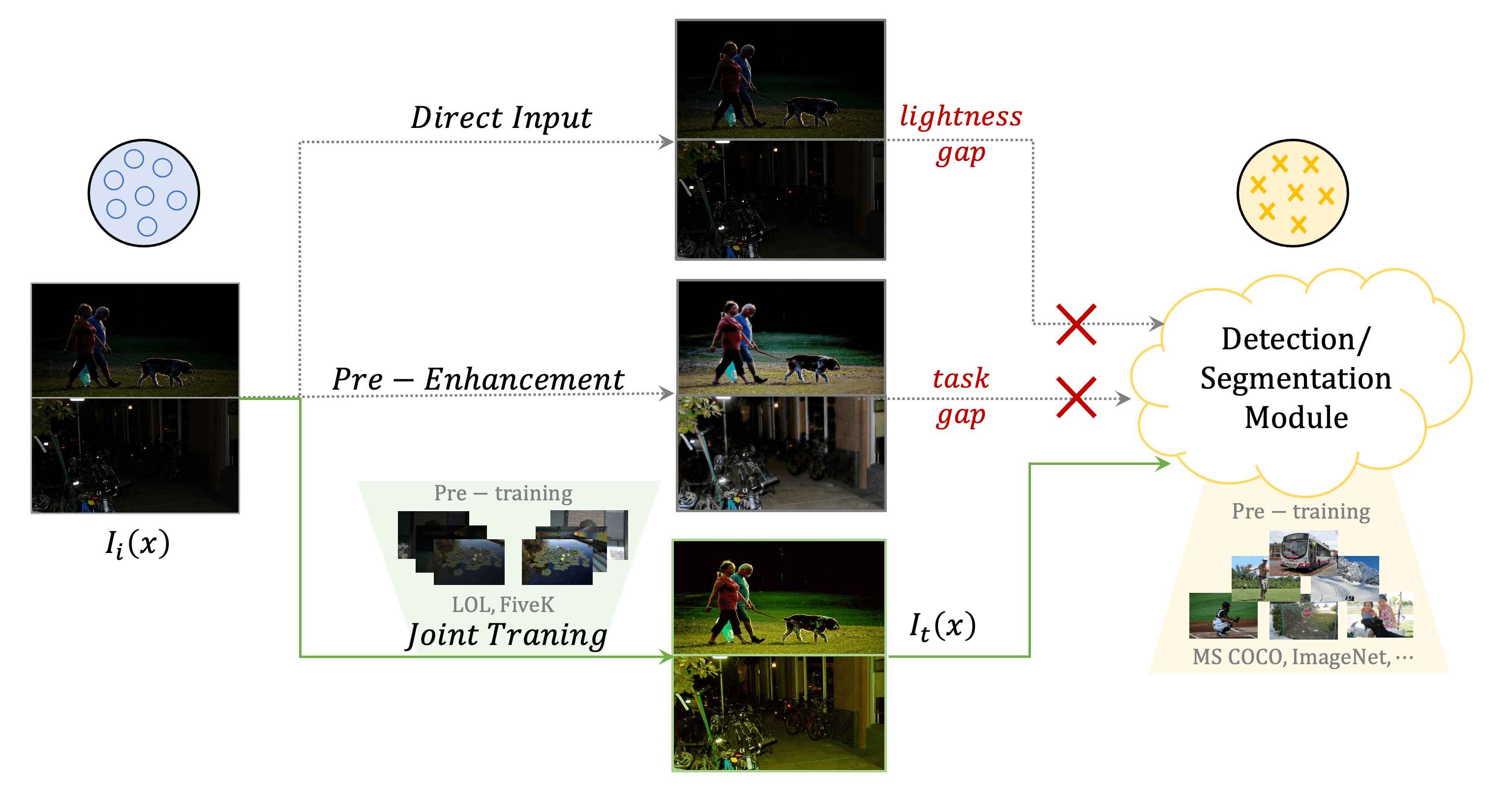}
    \caption{Joint training Enhancement Module with High-level Module.}
    \label{fig_high_level}
\end{figure}

For high-level vision tasks under challenging lighting conditions,shown in Fig.\ref{fig_high_level}, current high-level vision frameworks~\cite{DETR,yolov3,Deeplabv3plus} usually well-trained on large scale normal-light datasets (\textit{i.e.} MS COCO~\cite{coco_dataset}, ImageNet~\cite{imagenet}), so directly take low-light/ strong-light data as input would cause the lightness in-consistency, on the other hand, using image enhancement methods (Sec.4.3 in main text) to pre-process images may cause target inconsistency (human vision \textit{v.s.} machine vision)~\cite{ICCV_MAET}, since the goal of image restoration is image quality (\textit{i.e.} PSNR, SSIM) and the goal of detection/ segmentation is machine-vision accuracy (\textit{i.e.} mAP, mIOU).

An example is shown in Fig.~\ref{fig_high_level},
by attaching IAT to the downstream task module, our IAT could conduct object detection and semantic segmentation with the downstream frameworks. During training, we aim to minimise the downstream framework's loss function (\textit{i.e.} object detection loss $L_{obj}$ between detection prediction $\hat{t}$ and ground truth $t$) by jointly optimising the whole network's parameters (see Eq.~\ref{eq:high_level}).
Compared to the subsequent high-level module, the time-complexity and model storage of  our IAT main structure could be ignored (\textit{i.e.} IAT main structure \textit{v.s.} YOLO-V3~\cite{yolov3}, 417KB \textit{v.s.} 237MB).

\begin{equation}
\begin{aligned}
    \min\limits_{i \in \mathbb{I}, d\in \mathbb{D}} L_{obj}&( \hat{t}, t)\\
    I_t(x) = \mathbb{I}(I_1(x)), \quad & \hat{t} = \mathbb{D}(I_t(x))
\end{aligned}
\label{eq:high_level}
\end{equation}

We make the comparison experiments on low-light detection dataset EXDark~\cite{EXDark} and low-light semantic segmentation dataset ACDC~\cite{ACDC}. For object detection task we adopt the YOLO-V3~\cite{yolov3} object detector and for segmentation task we adopt DeepLabV3+~\cite{Deeplabv3plus} segmentation framework, the training and experiments' settings are follow the settings in Sec.4.3. 

Experimental results are shown in Table.~\ref{Tab:Comparision_high}, "original" means to take the original low-light images for training and evaluation, "pre-enhancement" means to pre-enhancement the EXDark~\cite{EXDark} and ACDC~\cite{ACDC} datasets with IAT model trained on LOL-V1 dataset~\cite{LOL_dataset} ("IAT (LOL)") and MIT-Adobe FiveK dataset~\cite{fivek_dataset} ("IAT (MIT5K)"). The "joint training" means to joint train IAT with the following high-level framework, and IAT model is separately random initialize ("IAT (none)"), initialize with LOL pre-train weights ("IAT (LOL)") and initialize with MIT-Adobe FiveK weights ("IAT (MIT5K)"), from Table.~\ref{Tab:Comparision_high} we could see that joint-training IAT with the high-level frameworks would further improve high-level visual performance, on both of object detection and semantic segmentation task.


\begin{table}[t]
\caption{Comparison experiments on low-light detection dataset EXDark~\cite{EXDark} and low-light semantic segmentation dataset ACDC~\cite{ACDC}.}
\centering
\begin{adjustbox}{max width = 0.95\linewidth}
\begin{tabular}{l|c|cc|ccc}
\Xhline{1.0pt}
\multirow{2}{*}{} & \multirow{2}{*}{original} & \multicolumn{2}{c|}{pre-enhancement} & \multicolumn{3}{c}{joint training}   \\ \cline{3-7} 
                  &                           & IAT (LOL)        & IAT (MIT5K)       & IAT (none) & IAT (MIT5K) & IAT (LOL) \\ \Xhline{0.6pt}
EXDark (mAP$\uparrow$)      & 76.4                      & 77.2             & 76.9              & 77.1       & 77.6        & \textbf{77.8}     \\ \hline
ACDC (mIOU$\uparrow$)       & 63.3                      & 62.1             & 61.3              & 61.5       &   62.1   &    \textbf{63.8}   \\ \Xhline{1.0pt}
\end{tabular}
\end{adjustbox}
\label{Tab:Comparision_high}
\vspace{-3mm}
\end{table}

\section{Ablation Studies}
\vspace{-3mm}
\noindent
\subsection{Contribution of each part.}
To evaluate each part's contribution in our IAT model, we make an ablation study on the low-light enhancement task of LOL-V2-real~\cite{LOL_dataset} dataset, and the low-light object detection task of EXDark~\cite{EXDark} dataset. We report the PSNR and SSIM results of the enhancement task and the mAP result of the detection task. We compare our normalization with LayerNorm~\cite{ln} and ResMLP's normalization~\cite{touvron2021resmlp}, and then evaluate different parts' contributions of the global branch (predict matrix and predict gamma value). The ablation results are shown in Table.~\ref{Ablition:part}. 

\begin{table}[t]
\caption{Experiments on LOL-V2-real~\cite{LOL_dataset} dataset (SSIM, PSNR) and EXDark~\cite{EXDark} dataset (mAP), shows each part's contribution of IAT.}

\centering
\setlength\tabcolsep{3pt}
\begin{adjustbox}{max width = 0.95\linewidth}
\begin{tabular}{cccccc|l|l|l}
\Xhline{1.0pt}
Local  & Layer & \cite{touvron2021resmlp}'s & Our & Global & Global & \multirow{2}*{PSNR$\uparrow$} & \multirow{2}*{SSIM$\uparrow$} & \multirow{2}*{mAP$\uparrow$}\\ 
Branch & Norm & Norm & Norm & (matrix) & (gamma) & ~ & ~ & ~ \\
\Xhline{0.6pt}
    $\surd$  &   $\surd$    &   &     &     &   &  18.80 & 0.762 & 75.8 \\  \hline
    $\surd$  &     &    $\surd$ &   &    &   & 19.61 (+0.81) &  0.776 (+0.014) & 75.8 (+0.0)\\  \hline
    $\surd$  &    &  &      $\surd$    &  &   &  20.01 (+1.21) &   0.786 (+0.024) & 76.3 (+0.5)\\  \hline
    $\surd$  &    &   &   $\surd$  &  $\surd$  &    &   21.95 (+3.15)   & 0.811 (+0.049)  & 76.5 (+0.7)\\ \hline
    $\surd$  &    &   &   $\surd$  &   &    $\surd$     &  22.76 (+3.96) &  0.805 (+0.043) & 76.7 (+0.9) \\ \hline
    $\surd$  &    &   &   $\surd$  & $\surd$  & $\surd$  &  \textbf{23.50} (+4.70)  & \textbf{0.824} (+0.062) &  \textbf{77.1} (+1.3) \\
\Xhline{1.0pt}
\end{tabular}
\end{adjustbox}
\label{Ablition:part}
\end{table}

\begin{table*}[t]
    \centering
    \begin{minipage}[t]{0.34\linewidth}
        \centering
        \caption{Blocks Number.
        }
        
        \setlength\tabcolsep{3.5pt}
        \resizebox{\textwidth}{!}{
        \begin{tabular}{c|c|c|c}
            \Xhline{0.8pt}
            \diagbox{$M$}{$A$} & 2     & 3     & 4     \\ 
            \Xhline{0.6pt}
            2  & 22.10 & 22.85 & 22.34 \\ \hline
            3  & 22.24 & \textbf{23.50} & 22.67 \\ \hline
            4  & 22.42 & 23.00 & 23.48 \\ 
            \Xhline{0.8pt}
        \end{tabular}
        }
        \label{Ablition:block}
    \end{minipage}
    \hspace{1mm}
    \begin{minipage}[t]{0.61\linewidth}
        \centering
        \caption{Channel Number.}
        
        \setlength\tabcolsep{4pt}
        \resizebox{\textwidth}{!}{
        \begin{tabular}{c|ccc}
            \Xhline{0.8pt}
            \multirow{2}{*}{\#Channel:\#Block} & \multirow{2}{*}{PSNR$\uparrow$} & \multirow{2}{*}{SSIM$\uparrow$}  & \#Param.$\downarrow$ \\
            ~ & ~ & ~ & (K) \\
            \Xhline{0.6pt}
             Long and Thin (12:4) & 22.60     & 0.807 & \textbf{86.22}          \\ \hline
             Short and Thick (24:2)   & 22.70     & 0.815 & 101.03         \\ \hline
            Ours (16:3)            & \textbf{23.50}     & \textbf{0.824} & 91.15          \\
            \Xhline{0.8pt}
        \end{tabular}
        }
        \label{Ablition:channel}
    \end{minipage}
\end{table*}

\vspace{2mm}
\noindent
\subsection{Blocks $\&$ Channels Ablation.}
To evaluate the scalability of our IAT model, we try the different block numbers and channel numbers in the local branch. We try different PEM numbers to generate $M$ and $A$. The PSNR results on LOL-V2-real~\cite{LOL_dataset} dataset has been shown in Table.\ref{Ablition:block}. It shows that keeping the same PEM number to generate $M$ and $A$ would be helpful to IAT's performance.

Keeping the same block number to generate $M$ and $A$, we then evaluate with similar parameters to answer whether the local branch should be ``short and thick" or ``long and thin". The local branch's block number and channel number are respectively set to $2/24$ and $4/12$ for comparison. The results of PSNR, SSIM and model parameters are reported in Table.~\ref{Ablition:channel}.

\section{Additional Qualitative Results.}
\setcounter{table}{0}
\setcounter{figure}{0}

In this section we show more qualitative results on low-level vision tasks: image enhancement (LOL (V1 $\&$ V2-real)~\cite{LOL_dataset}, MIT-Adobe FiveK~\cite{fivek_dataset}) and exposure correction~\cite{Exposure_2021_CVPR}.

\subsection{Image Enhancement Results}

\begin{figure}[t]
    \centering
    \includegraphics[width=1.0\linewidth]{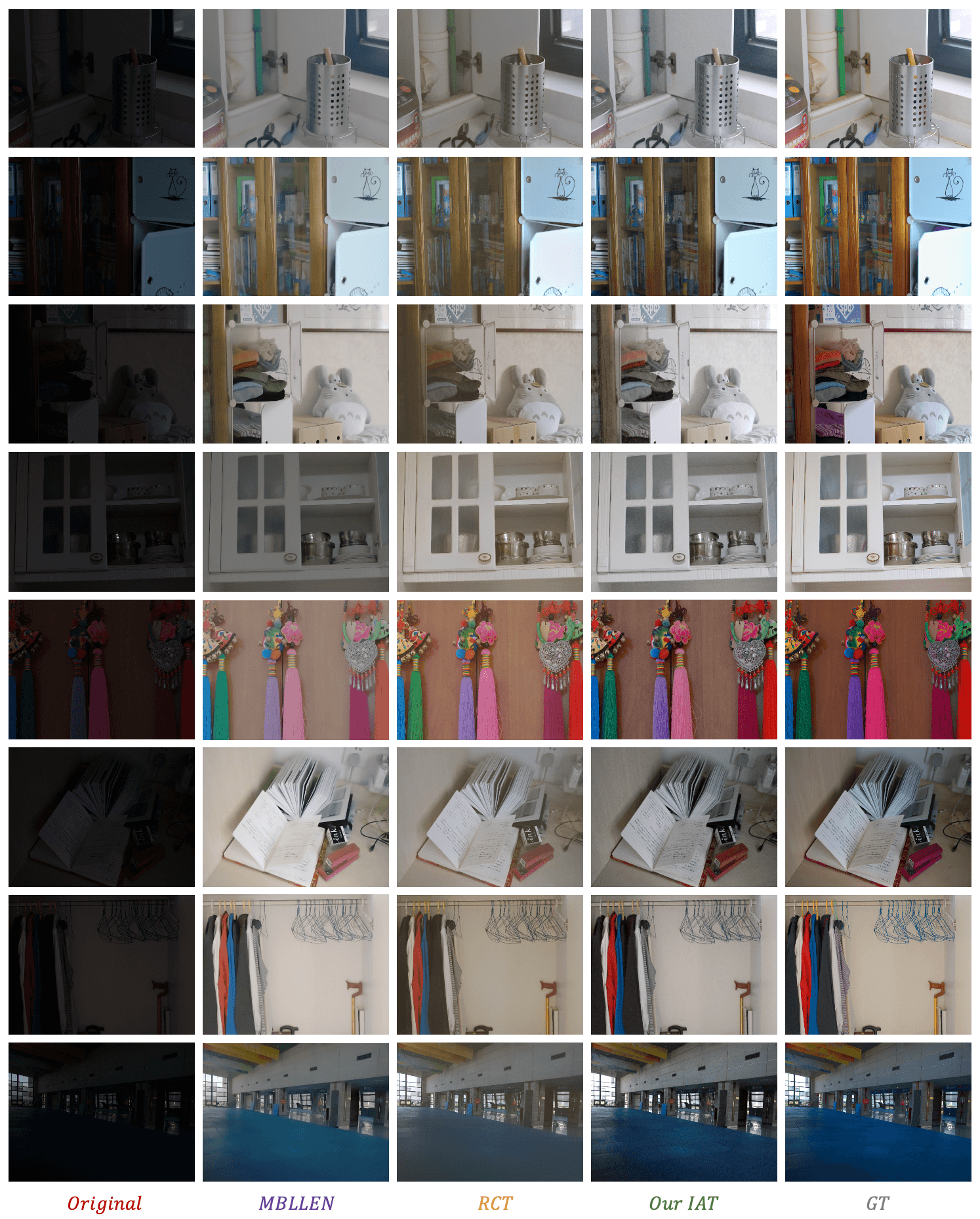}
    \caption{Qualitative comparison results on  LOL-V1~\cite{LOL_dataset} dataset, compare with enhancement methods MBLLEN~\cite{Lv2018MBLLEN} and RCT~\cite{RCT_ICCV21}.}
    \label{fig:LOL_V1}
\end{figure}

\begin{figure}[t]
    \centering
    \includegraphics[width=1.0\linewidth]{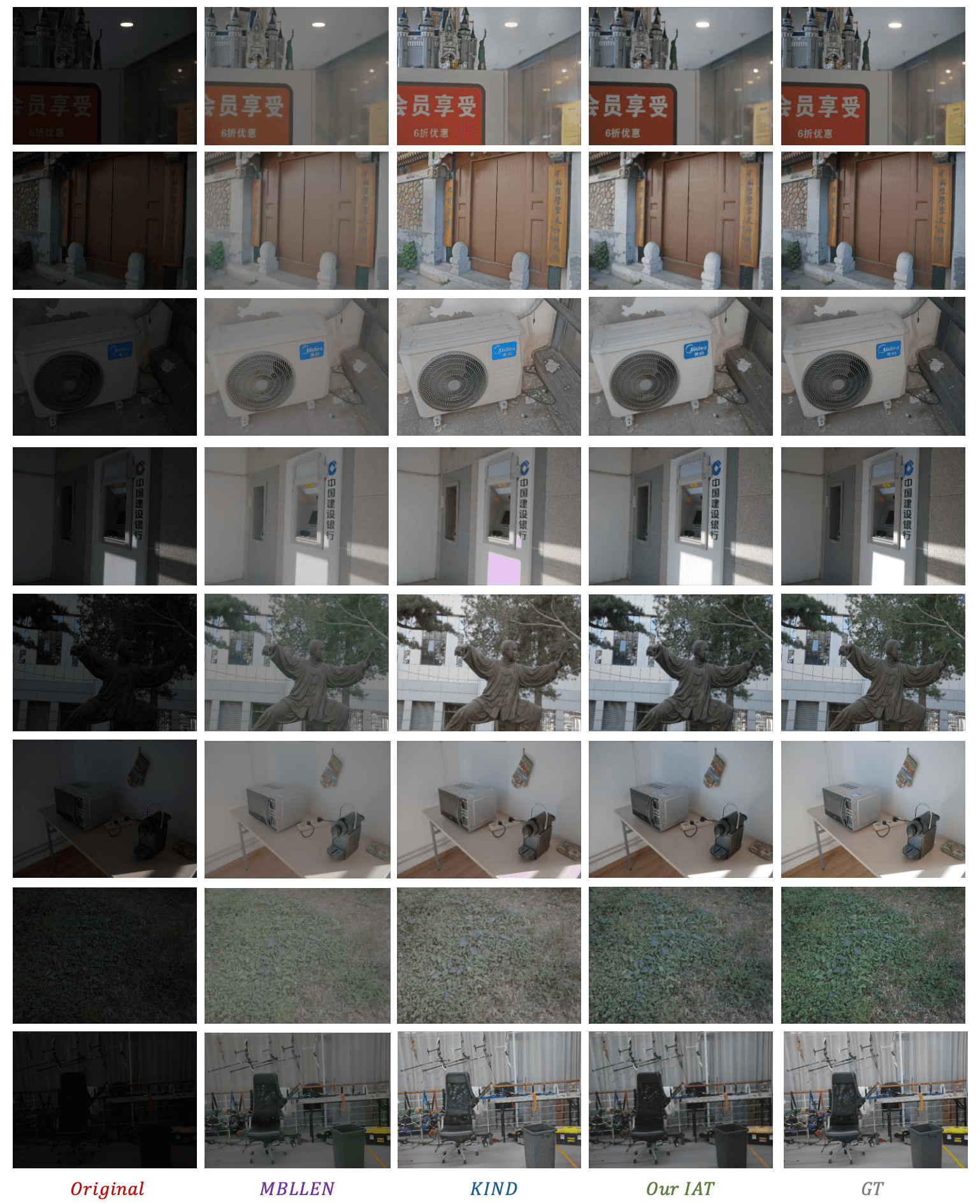}
    \caption{Qualitative comparison results on  LOL-V2-real~\cite{LOL_dataset} dataset, compare with enhancement methods MBLLEN~\cite{Lv2018MBLLEN} and KIND~\cite{KIND}.}
    \label{fig:LOL_V2}
\end{figure}

\begin{figure}[t]
    \centering
    \includegraphics[width=1.0\linewidth]{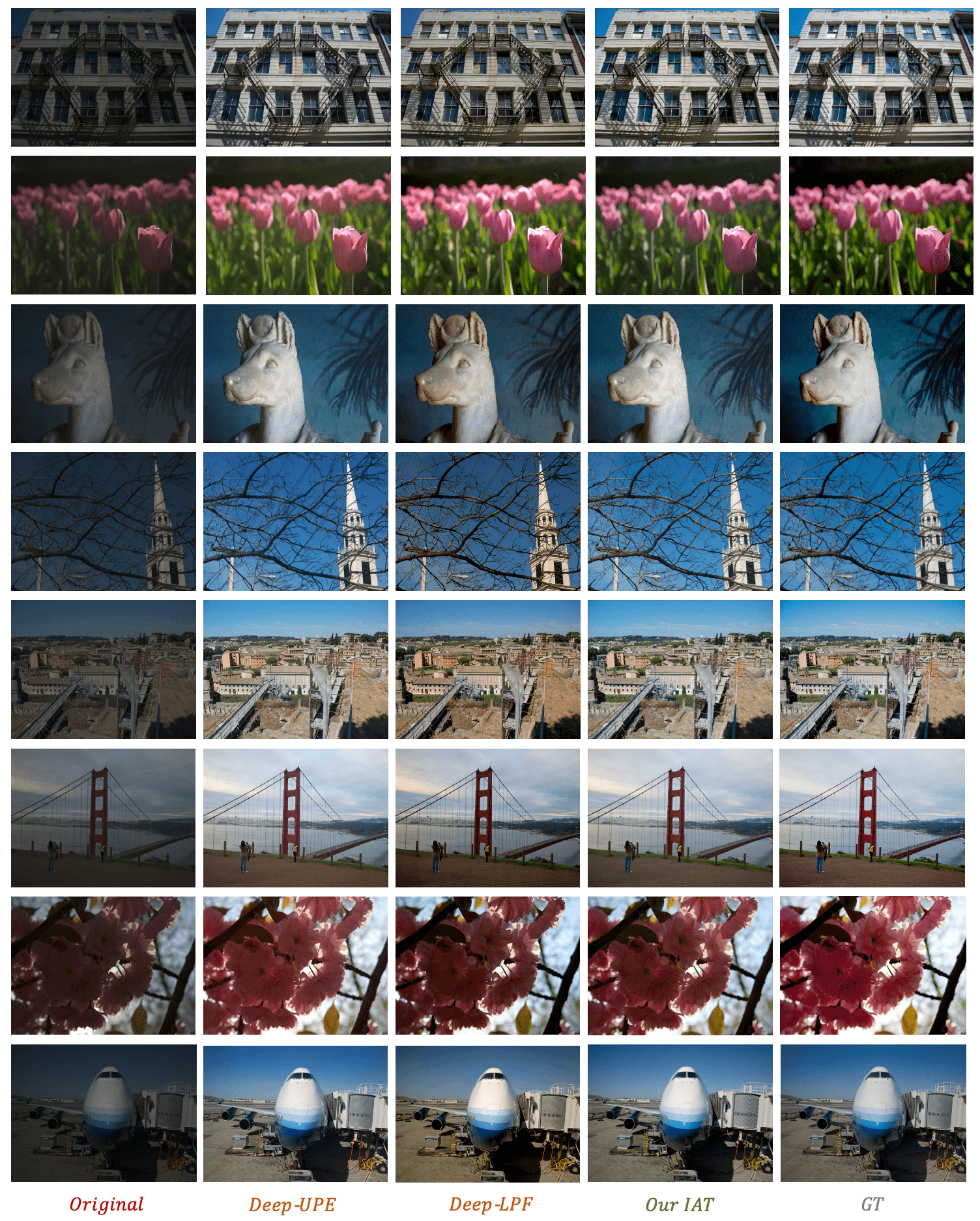}
    \caption{Qualitative comparison results on MIT-Adobe FiveK~\cite{fivek_dataset} dataset, compare with enhancement methods Deep-UPE~\cite{DeepUPE_2019_CVPR} and Deep-LPF~\cite{Deep_LPF}.}
    \label{fig:MIT5k}
\end{figure}

Fig.~\ref{fig:LOL_V1} shows the image enhancement results on LOL-V1 dataset~\cite{LOL_dataset} compare with RCT~\cite{RCT_ICCV21} and MBLLEN~\cite{Lv2018MBLLEN}, Fig.~\ref{fig:LOL_V2} shows the image enhancement results on LOL-V2-real dataset~\cite{LOL_dataset} compare with MBLLEN~\cite{Lv2018MBLLEN} and KIND~\cite{KIND}. Fig.~\ref{fig:MIT5k} shows the image enhancement results on MIT-Adobe FiveK dataset~\cite{fivek_dataset} compare with Deep-UPE~\cite{DeepUPE_2019_CVPR} and Deep-LPF~\cite{Deep_LPF}. We could see that IAT can generate higher quality images which closer to reference target image $I_t$. Meanwhile IAT also take much fewer parameters and less inference time. 

\subsection{Exposure Correction Results}
\begin{figure}[t]
    \centering
    \includegraphics[width=1.0\linewidth]{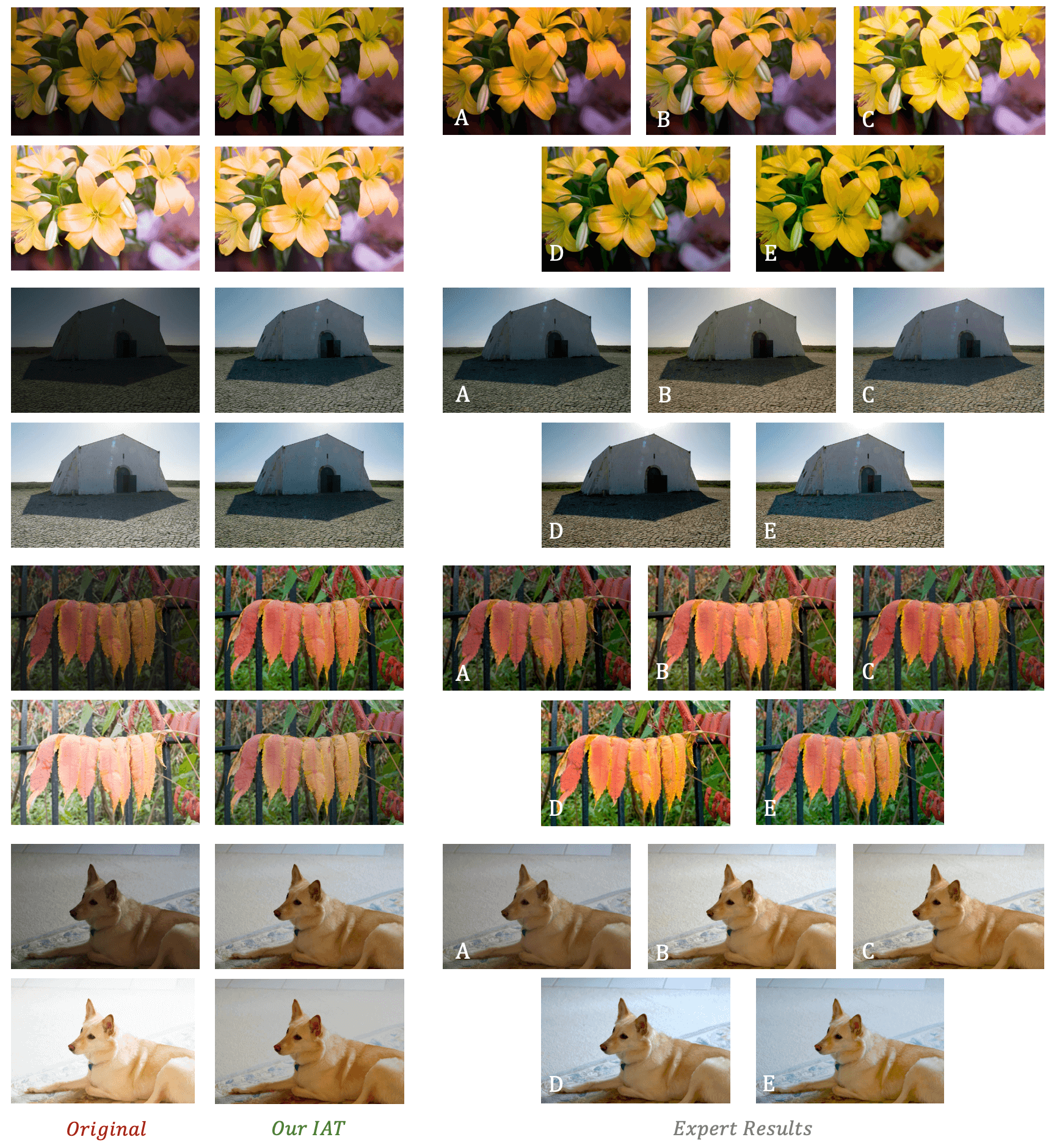}
    \caption{Qualitative comparison results of both under-exposure and over-exposure images on exposure correction dataset~\cite{Exposure_2021_CVPR}, left is input image, second row is output of our IAT, right are 5 experts' results.}
    \label{fig:exposure}
\end{figure}

Fig.~\ref{fig:exposure} shows the exposure correction results on~\cite{Exposure_2021_CVPR} dataset, we show both under-exposure and over-exposure results of our IAT, and compare to five experts' results. IAT also generate high quality images, and have ability to handle under/over-exposure at same time.

\end{document}